\documentclass[10pt,twocolumn,letterpaper,table]{article}
\makeatletter
\@namedef{ver@everyshi.sty}{}
\makeatother
\usepackage[pagenumbers]{cvpr} % To force page numbers, e.g. for an arXiv version

\usepackage[pdftex]{graphicx}
\usepackage{amsmath}
\usepackage{amssymb}
\usepackage{booktabs}

\usepackage[utf8]{inputenc} % allow utf-8 input
\usepackage{url}            % simple URL typesetting
\usepackage{booktabs}       % professional-quality tables
\usepackage{amsfonts}       % blackboard math symbols
\usepackage{nicefrac}       % compact symbols for 1/2, etc.
\usepackage{microtype}      % microtypography
\usepackage{xspace}
\usepackage{color}
\usepackage[dvipsnames,table,xcdraw]{xcolor}
\usepackage{color, colortbl}
\usepackage[misc,geometry]{ifsym} % for \Letter
\definecolor{tabhighlight}{HTML}{e5e5e5}
\definecolor{citecolor}{HTML}{0071bc}

\usepackage{amsmath}
\usepackage{amsfonts}
\usepackage{amssymb}
\usepackage{wrapfig}
\usepackage{subcaption}
\usepackage{multirow}
\usepackage{mathtools} 
\usepackage{bm}

\usepackage{verbatim}

\usepackage{anyfontsize}

\usepackage{microtype}
\usepackage{graphicx}
\usepackage{booktabs} % for professional tables
\usepackage{multirow}
\usepackage{amsmath,amssymb}
\usepackage{booktabs}
\usepackage{caption,subcaption}

\usepackage{xcolor}
\definecolor{mygreen}{HTML}{3cb44b}
\definecolor{skyblue}{HTML}{beffff}
\definecolor{lightgreen}{HTML}{90ee90}

\usepackage{tcolorbox}
\usepackage{enumitem}
\setitemize{itemsep=10pt,topsep=0pt,parsep=0pt,partopsep=0pt}
\usepackage{adjustbox}

\newcommand{\RN}[1]{%
    \textup{\lowercase\expandafter{\it \romannumeral#1}}%
}
\usepackage{tabu}
\newcommand{\beq}{\vspace{0mm}\begin{equation}}
\newcommand{\eeq}{\vspace{0mm}\end{equation}}
\newcommand{\beqs}{\vspace{0mm}\begin{eqnarray}}
\newcommand{\eeqs}{\vspace{0mm}\end{eqnarray}}
\newcommand{\barr}{\begin{array}}
\newcommand{\earr}{\end{array}}

\newcommand{\vv}{\boldsymbol{v}}

\usepackage{color, colortbl}
\definecolor{Gray}{gray}{0.93}

\usepackage{lipsum}

\usepackage{xcolor,amsmath}
\usepackage[linesnumbered,ruled,vlined]{algorithm2e}
\DontPrintSemicolon

\usepackage{color, colortbl}

\definecolor{emerald}{rgb}{0.31, 0.78, 0.37}

\newcommand{\MyColorBox}[2][red]%
{%
    \settowidth{\Width}{#2}%
    \colorbox{#1}%
    {%      
        \raisebox{-\DepthReference}%
        {%
                \parbox[b][\HeightReference+\DepthReference][c]{\Width}{\centering#2}%
        }%
    }%
}

\definecolor{codegray}{gray}{0.9}

\SetKwComment{Comment}{\color{green!50!black}\# }{}

\SetKwProg{Function}{def}{:}{}

\SetKwProg{For}{for}{:}{}
\SetKwProg{If}{if}{:}{}

\def\Secref#1{Section~\ref{#1}}
\def\eqref#1{equation~\ref{#1}}
\def\1{\bm{1}}

\def\vp{{\bm{p}}}
\def\vq{{\bm{q}}}

\def\vu{{\bm{u}}}
\def\vv{{\bm{v}}}

\def\vx{{\bm{x}}}
\def\vy{{\bm{y}}}
\def\vz{{\bm{z}}}

\def\mP{{\bm{P}}}
\def\mQ{{\bm{Q}}}

\def\mU{{\bm{U}}}
\def\mV{{\mathbf{V}}}
\def\mW{{\mathbf{W}}}
\DeclareMathAlphabet{\mathsfit}{\encodingdefault}{\sfdefault}{m}{sl}
\SetMathAlphabet{\mathsfit}{bold}{\encodingdefault}{\sfdefault}{bx}{n}

\newcommand{\tableCellHeight}{1}
\newcommand{\tabstyle}[1]{
  \setlength{\tabcolsep}{#1}
  \renewcommand{\arraystretch}{\tableCellHeight}
  \centering
  \small
}
\newcommand{\ours}{{MVLPT}}
\newcommand{\vlpt}{{UPT}\xspace}
\newcommand{\elevater}{\textsc{Elevater}\xspace}
\newcommand{\romannum}[1]{\romannumeral #1} % roman numbering
\newcommand{\rotbox}[1]{\rotatebox{90}{#1}}
\newcommand{\hgreen}[1]{\textcolor{ForestGreen}{\textbf{#1}}} % highlight color
\newcommand{\hblue}[1]{\textcolor{NavyBlue}{\textbf{#1}}} % highlight color
\usepackage[pagebackref,breaklinks,colorlinks]{hyperref}

\usepackage[capitalize]{cleveref}
\crefname{section}{Sec.}{Secs.}
\Crefname{section}{Section}{Sections}
\Crefname{table}{Table}{Tables}
\crefname{table}{Tab.}{Tabs.}

\def\cvprPaperID{} % *** Enter the CVPR Paper ID here
\def\confName{CVPR}
\def\confYear{2023}

\begin{document}

%%%%%%%%% TITLE - PLEASE UPDATE
\title{Multitask Vision-Language Prompt Tuning}

\author{ 
  Sheng Shen$^\dagger$\thanks{Equal contribution} \,\, Shijia Yang$^{\dagger*}$ \,\, Tianjun Zhang$^{\dagger*}$ \,\, Bohan Zhai$^\ddag$ \\ 
  {Joseph E. Gonzalez}$^\dagger$ \,\, {Kurt Keutzer}$^\dagger$ \,\, {Trevor Darrell}$^\dagger$ \\
  $^\dagger$University of California, Berkeley, $^\ddag$ByteDance Inc. \\
  \texttt{\small \{sheng.s,shijiayang,tianjunz,jegonzal,keutzer,trevordarrell\}@berkeley.edu} \\
}
% \autho
% \author{
% For a paper whose authors are all at the same institution,
% omit the following lines up until the closing ``}''.
% Additional authors and addresses can be added with ``\and'',
% just like the second author.
% To save space, use either the email address or home page, not both
% \and
% Second Author\\
% Institution2\\
% First line of institution2 address\\
% {\tt\small secondauthor@i2.org}
% }
\maketitle

%%%%%%%%% ABSTRACT
\begin{abstract}
Prompt Tuning, conditioning on task-specific learned prompt vectors,  has emerged as a data-efficient and parameter-efficient method for adapting large pretrained vision-language models to multiple downstream tasks.
However, existing approaches usually consider learning prompt vectors for each task independently from scratch, thereby failing to exploit the rich shareable knowledge across different vision-language tasks.
In this paper, we propose multitask vision-language prompt tuning (\ours{}), 
which incorporates cross-task knowledge into prompt tuning for vision-language models. 
Specifically, 
%we explore this cross-task knowledge in two ways:
(i) %multitask prompt initialization: 
we demonstrate the effectiveness of learning a single transferable prompt from multiple source tasks to initialize the prompt for each target task;
(ii) %multitask prompt adaption: 
we show many target tasks can benefit each other from sharing prompt vectors and thus can be jointly learned via multitask prompt tuning. 
We benchmark the proposed \ours{} using three representative prompt tuning methods, namely text prompt tuning, visual prompt tuning, and the unified vision-language prompt tuning. 
Results in 20 vision tasks demonstrate that the proposed approach outperforms all single-task baseline prompt tuning methods, setting the new state-of-the-art on the few-shot \elevater benchmarks and cross-task generalization benchmarks. 
% To understand the cross-task knowledge further, 
To understand where the cross-task knowledge is most effective, we also conduct a large-scale study on task transferability with 20 vision tasks in 400 combinations for each prompt tuning method.
It shows that the most performant \ours{} for each prompt tuning method prefers different task combinations and many tasks can benefit each other, depending on their visual similarity and label similarity.  
Code is available at \url{https://github.com/sIncerass/MVLPT}. 
\end{abstract}

\maketitle

\section{Introduction}
\label{sec:intro}
\begin{figure}[h]
% \begin{minipage}{\textwidth} 
% \begin{center}
    \includegraphics[width=\linewidth]{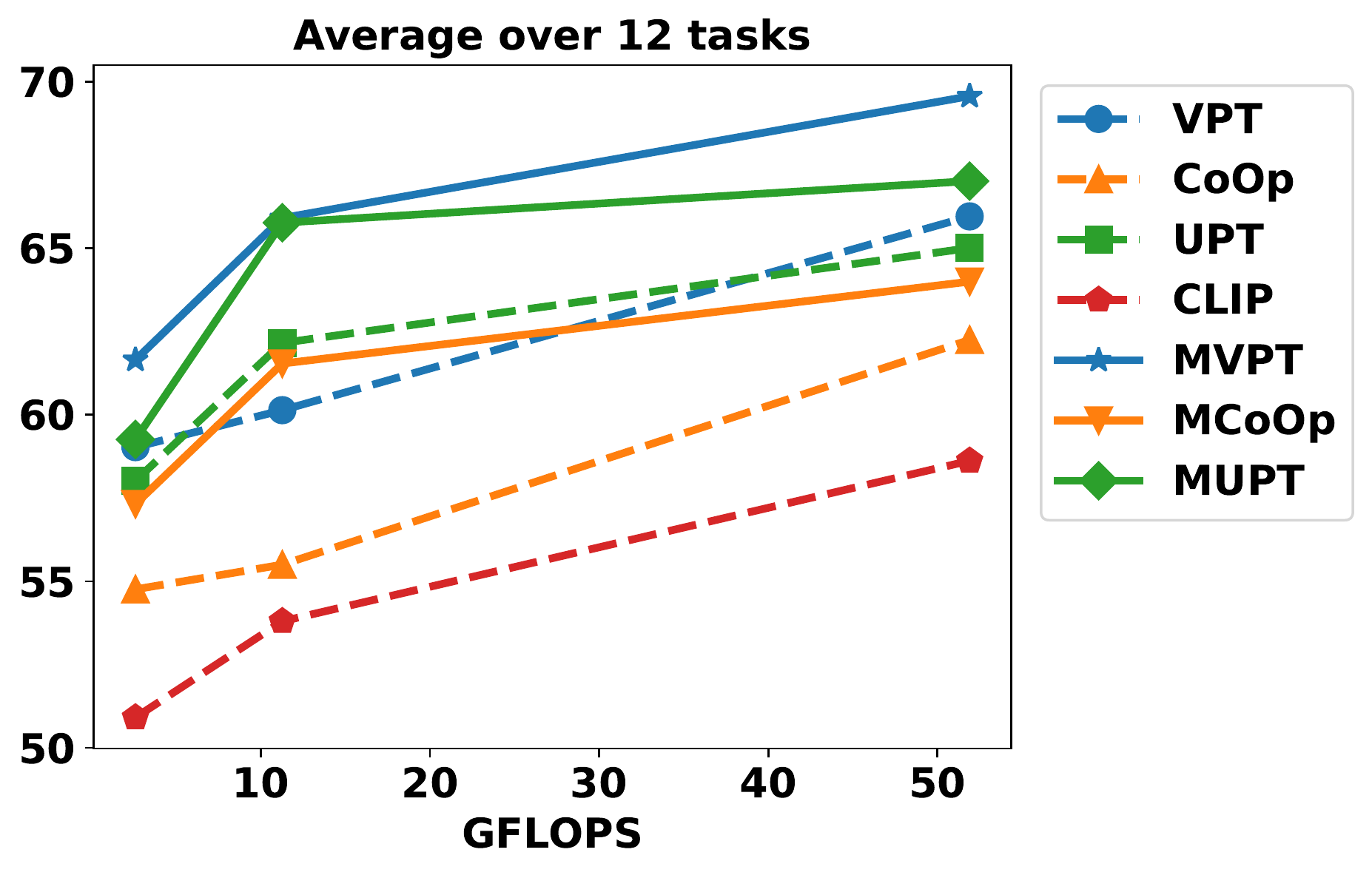}
    % \includegraphics[width=0.74\textwidth]{figs/arch.pdf}
% \end{center}
% \vspace{-0.1cm}
% \end{minipage}
% \hspace{0.01\textwidth}
% \begin{minipage}{0.24\textwidth} 
% \centering
\caption{
\footnotesize
Our \ours{} approach (MCoOp, MVPT, MUPT)
---which transfers a prompt learned from a mixture of source tasks (here, 11 Image Classification tasks) onto non-overlapped target tasks---
outperforms vanilla CoOp~\cite{zhou2021coop}, VPT~\cite{jia2022visual} on 12 \elevater tasks by a large margin, across all CLIP model sizes (ViT-B/32, ViT-B/16 and ViT-L/14). 
% See Section \ref{sec:ablation} for full details. 
% unified prompt $\mU$ that is applied to (\textbf{b}) CLIP \emph{text} encoder and (\textbf{c}) CLIP \emph{image} encoder.
}
    \label{fig:mvlpt_scaling}
% \end{minipage}
\end{figure}

\begin{figure*}[t!]
\centering
\includegraphics[width=0.74\textwidth]{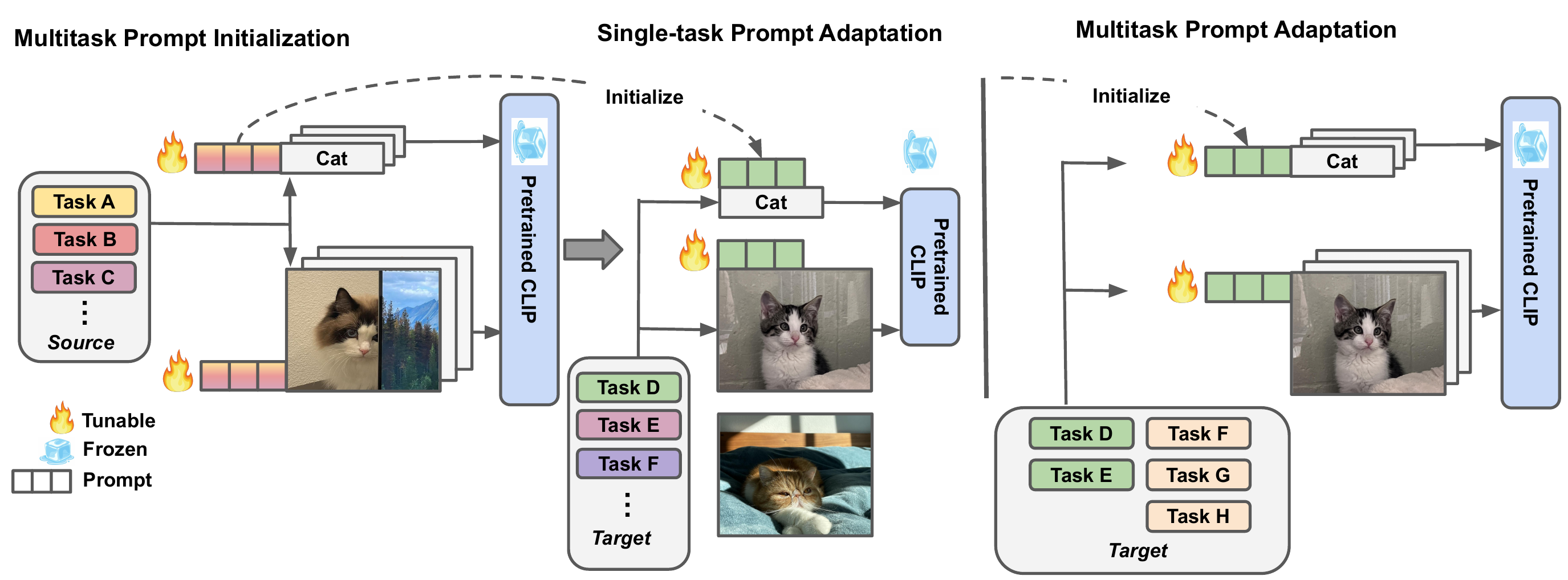}
\caption{An illustration of our \textit{multitask prompt initialization} (left) and \textit{multitask prompt adaptation} (right)  approaches in \ours{}. 
\textbf{Left}: We learn single generic source prompt vector on various \emph{source tasks}, which is then used to initialize the prompt for each single \emph{target task}. 
\textbf{Right}: After use the source prompt vector for initialization. We group relevant \emph{target tasks} together and perform multitask prompt tuning within each group. Noted that grouping one task means single-task adaptation. 
(see \Secref{sec:method} for details).
}
\vspace{-2mm}
\label{fig:approach}
\end{figure*}

Recent large-scale vision-language models, pretrained on a wide variety of images with natural language supervision (\ie, CLIP~\cite{radford2021learning}, ALIGN~\cite{jia2021scaling} and Florence~\cite{yuan2021florence}), have demonstrated strong open-set recognition abilities for image classification in-the-wild~\cite{radford2021learning,li2022elevater} and open-vocabulary detection~\cite{gu2021open}. 
Despite the impressive zero-shot transfer capabilities, adapting these large-scale vision-language models to downstream tasks presents its own challenges. 
It is usually prohibitive to fine-tune the entire model due to both huge parameter sizes and well-known overfitting issues for few-shot learning. 

Such a trend emerges the essential need to study different adaptation methods~\cite{houlsby2019parameter,hu2021lora,li2021prefix}, where Prompt Tuning~\cite{lester2021power,zhou2021coop} has shown to be one of the most effective strategies. 
Typically, Prompt Tuning tunes only a small number of parameters for each task in a model’s input spaces (prompt vectors) while keeping the pretrained model frozen. 
It was first introduced in NLP community~\cite{lester2021power,li2021prefix,liu2021gpt} and has recently demonstrated superior few-shot adaptation performance~\cite{zhou2021coop,zhou2022conditional,jia2022visual} for vision-language models. 
CoOp~\cite{zhou2021coop} and VPT~\cite{jia2022visual} are two representative vision-language prompt tuning methods, in which the former uses a textual prompt and the latter leverages the visual prompt. 

% Problem of single task adaptation and multi task tuning in language
However, on the one hand, most of these contemporary vision-language prompt tuning methods (\ie, CoOp, VPT) focuses on learning a prompt for each downstream task independently, failing to incorporate cross-task knowledge when adapting to various downstream tasks. 
On the other hand, multitask learning has a rich literature~\cite{thrun1995learning,caruana1997multitask,zhang2021survey,standley2020tasks} for vision. 
Applying multitask prompt tuning to language models has also presented impressive few-shot~\cite{liu2022few,asai2022attentional} or zero-shot generalization capability~\cite{chung2022scaling,sanh2021multitask}. 
This motivates us to investigate the question: \emph{Can vision-language model also benefit from multitask knowledge sharing via prompt tuning during adaptation?}
% Our proposed method

To this end, we propose multitask vision language prompt tuning (\ours{}), to the best of our knowledge, the first method incorporating the cross-task knowledge into vision-language prompt tuning. 
\ours{} is a simple yet effective way to enable information sharing between multiple tasks. \ours{} consists of two stages: \emph{multitask source prompt initialization} and \emph{multitask target prompt adaptation}. 
% We explore the two facets: (i) multitask prompt initialization and (ii) multitask prompt adaptation. 
Specifically, multitask prompt initialization first learns shared prompt vectors from various source tasks. 
Then this shareable prompt can be used to initialize the prompt for target tasks. 
To adapt to target tasks, multitask prompt adaptation will group relevant tasks together then perform multitask prompt tuning within the selected groups. 
We remark that we could also perform single-task adaption with setting group size as one. 
This simple scheme enables passing cross-task knowledge from \emph{source tasks} to \emph{target tasks} through multitask prompt initialization, and exploiting shareable knowledge within \emph{target tasks} via multitask prompt adaptation further.

We conduct extensive evaluations of \ours{} on 20 vision tasks in few-shot \elevater~\cite{li2022elevater} in Section~\ref{subsec:multitask_prompt_adapt}.
Comparing to CoOp~\cite{zhou2021coop}, VPT~\cite{jia2022visual} and \vlpt (Section~\ref{subsec:preliminaries}), \ours{} improves the baselines by 0.72\%, 1.73\% and 0.99\% respectively and sets the new state-of-the-art on 20-shot \elevater benchmark. 
We also show the strong generalizability of \ours{} where \ours{} improves CoOp, VPT and \vlpt by 1.73\%, 4.75\% and 4.53\%, respectively on cross-task generalization benchmark in Section~\ref{subsec:multitask_prompt_init} and study task transferability with the 20 vision tasks and in 400 combinations for each prompt method in Section~\ref{subsec:multitask_task_transfer}. 

In summary, we make the following contributions:
\begin{itemize}
    \item We propose the multitask vision-language prompt tuning (\ours{}) framework, including multitask prompt initialization and multitask prompt adaptation, and demonstrate the efficacy for each component.  
    \item We rigorously study the task transferability across 20 vision tasks with 400 combinations for each prompt tuning method to understand when \ours{} is most effective. 
    \item We systematically evaluate the proposed \ours{} on the few-shot \elevater and cross-task generalization benchmarks, which sets the new state-of-the-art on 20-shot \elevater benchmark. 
\end{itemize}

\section{Related Work}
\label{sec:related_work}
\begin{figure*}[ht]
% \begin{minipage}{0.74\textwidth} 
\begin{center}
    \includegraphics[width=0.74\textwidth]{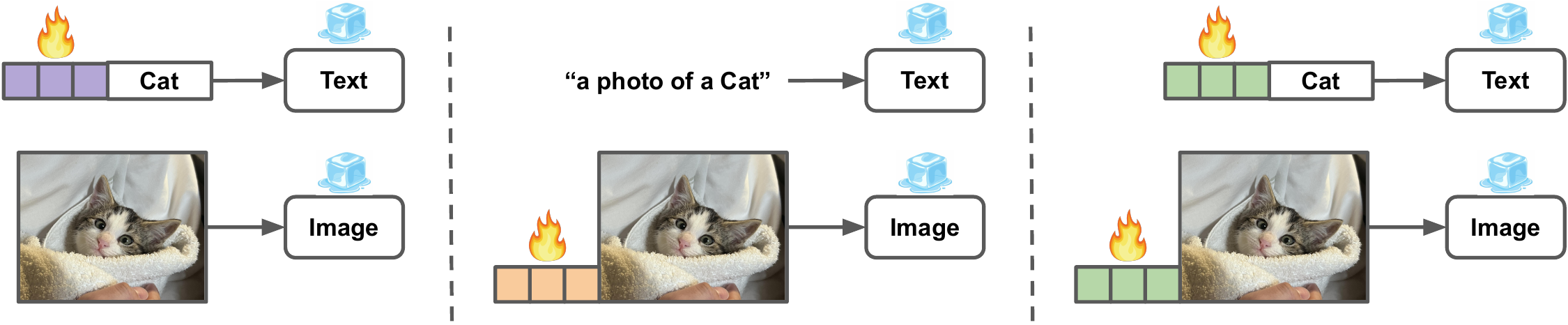}
\end{center}
% \end{minipage}\hspace{0.01\textwidth}
% \begin{minipage}{0.24\textwidth} 
\centering
\caption{The architecture of (\textbf{a}) CoOp (textual prompt tuning), (\textbf{b}) VPT (visual prompt tuning), and (\textbf{c}) \vlpt (unified prompt tuning). 
}
    \label{fig:different_prompt_method}
% \end{minipage}
\end{figure*}

\paragraph{Vision-Language Models.}
Vision-Language models targeted at aligning images and texts into a joint embedding space~\cite{radford2021learning,jia2021scaling,zhang2020contrastive,chen2022pali}. 
The resulting vision-language model~\cite{gan2022vision} typically has three key components: image, text encoding and the design of loss functions for alignment. 
Traditionally, models are often designed and learned independently for images and texts modules, with an extra module (losses) to connect these two outputs. 
For instance, images can be encoded using hand-crafted descriptors~\cite{socher2013zero,elhoseiny2013write} or neural networks~\cite{frome2013devise,lei2015predicting}. 
Texts can be encoded with pre-trained word vectors~\cite{socher2013zero,frome2013devise}, the frequency-based  features~\cite{elhoseiny2013write,lei2015predicting}. 
To align these two modality, metric learning~\cite{frome2013devise}, multi-label classification~\cite{joulin2016learning,gomez2017self}, and n-gram language learning~\cite{li2017learning} are widely adopted.

Recently, with the rise of large-scale pretraining, vision-language models~\cite{li2019visualbert,tan2019lxmert,su2019vl,li2020oscar,li2021align,kim2021vilt,zhang2021vinvl,radford2021learning,jia2021scaling,furst2021cloob,yao2021filip,li2021supervision,shen2021much,wang2021simvlm,wang2022image,alayrac2022flamingo,li2022blip,shen2022k,yu2022coca} bridge the two modalities by learning two encoders jointly. 
Also, the models are usually built with much larger neural networks (up to 80B parameters as in~\cite{alayrac2022flamingo}) and larger dataset. 
As discussed in Zhe et al.~\cite{gan2022vision}, recent successes in vision-language models can mainly attribute to the developments in \romannum{1}) Transformers~\cite{vaswani2017attention}, \romannum{2}) contrastive representation learning~\cite{chen2020simple,he2020momentum,henaff2020data,yu2022coca}, and \romannum{3}) web-scale training datasets~\cite{radford2021learning,jia2021scaling,yuan2021florence}. 
A representative approach is CLIP~\cite{radford2021learning}, which trains two neural network-based encoders using a contrastive loss to match image-text pairs. 
After consuming 400 million data pairs, the CLIP model demonstrates a remarkable zero-shot image recognition capability. 

\paragraph{Prompt Tuning.}
This topic originates from the NLP community~\cite{lester2021power,li2021prefix,liu2021gpt} to improve practical applicability of large-scale pre-trained language models~\cite{devlin2019bert,radford2019language,zhang2022opt,brown2020language,scao2022bloom}.  
Concretely, the target NLP task will be reformulated as a ``fill-in-the-blank'' cloze test, which queries the language model to predict the masked token in ``\texttt{I enjoyed the movie. \underline{It was} [MASK].}'' as either ``positive'' or ``negative'' for sentiment classification. 
The vital component lies in both designing the ``verbalizer'' (the label for the mask token) and the underlined part, known as prompt (template), in such a format familiar to the model. 
Several efforts have since focused on developing prompt-based learning approaches with carefully handcrafted prompts~\cite{schick2020s}, prompt mining and paraphrasing~\cite{jiang2020can}, gradient-based search for improved prompts~\cite{shin2020autoprompt}, and automatic prompt generation~\cite{gao2020making}. 
The use of hard prompts, however, was found to be sub-optimal and sensitive to the choice of the prompt~\cite{zhao2021calibrate,liu2021gpt}.
As such, more recent work has shifted toward learning continuous prompt learning methods~\cite{zhong2021factual,li2021prefix,lester2021power,liu2021gpt,liu2022p,vu2022spot}, where the main idea is to turn a prompt into a set of continuous vectors that can be end-to-end optimized with respect to an objective function, which is also is most related to our research. 
See Liu et al.~\cite{liu2021pre} for a more comprehensive survey. 
More recently, \cite{kojima2022large, wei2022chain, zhou2022least} focus on writing out reasoning steps explicitly in the ``chain-of-thoughts'' prompt. 

In computer vision, prompt learning is a nascent research direction that has only been explored very recently~\cite{zhou2021coop,yao2021cpt,rao2022denseclip,ju2021prompting,zhang2021pointclip,zhou2022conditional,jia2022visual,bar2022visual}.
Noticed that all of previous vision-language prompt studies focus on the single task prompt learning while we study incorporating cross-task knowledge in the prompt tuning process. 

\paragraph{Multi-task prompt tuning} Very close to our proposal, multitask prompt tuning has recently been explored extensively in the NLP community. 
Specifically, one line of the research focuses on multitask prompt finetuning  \cite{wei2021finetuned,sanh2021multitask,wang2022benchmarking,muennighoff2022crosslingual,chung2022scaling} that further finetunes the pretrained model on massive, human-crafted, (thousands of) prompt-formatted downstream tasks and find the resulting finetuned model expresses strong generalization ablity to unseen NLP tasks.
Another line of the research explores 
multitask continuous prompt tuning \cite{vu2022spot,asai2022attentional}.
This is very similar to our setting except that they only focus on multitask prompt initialization and NLP tasks. 

\section{Methodology}
\label{sec:method}
\begin{figure*}[t!]
\centering
\begin{subfigure}{.33\textwidth}\includegraphics[width=\textwidth]{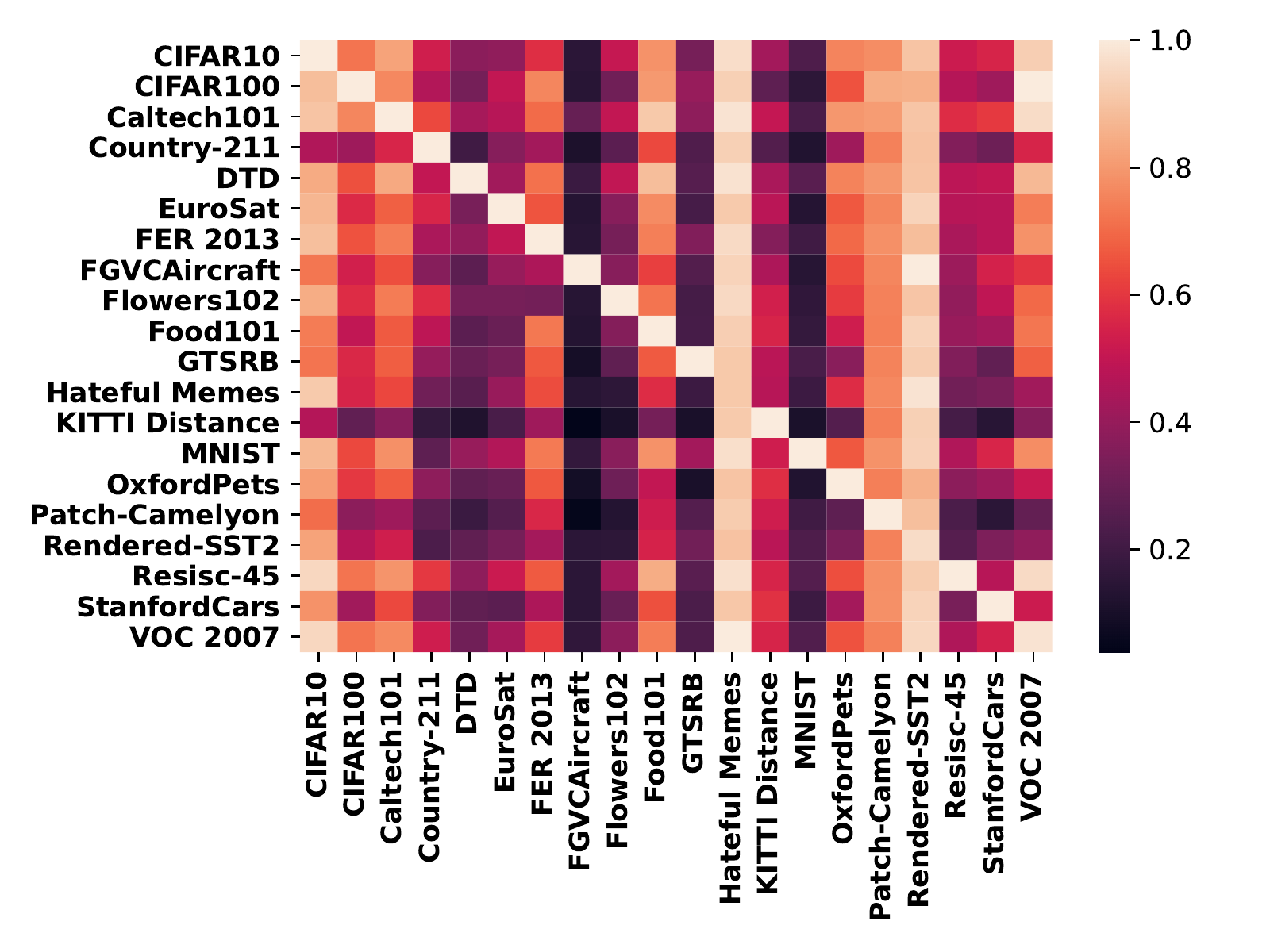}
\caption{CoOp Transferability.}
\end{subfigure}
\begin{subfigure}{.33\textwidth}\includegraphics[width=\textwidth]{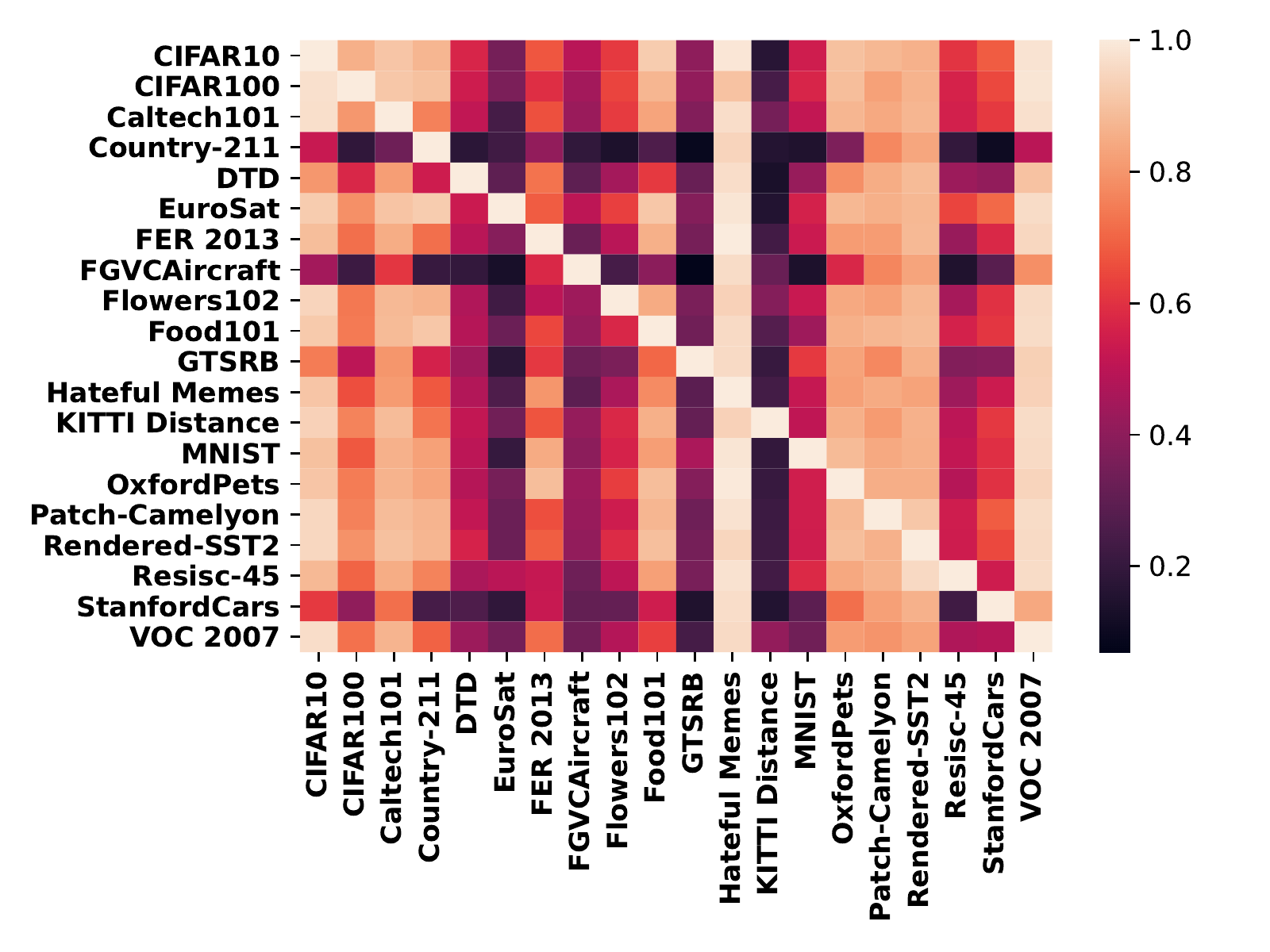}
\caption{VPT Transferability.}
\end{subfigure}
\begin{subfigure}{.33\textwidth}\includegraphics[width=\textwidth]{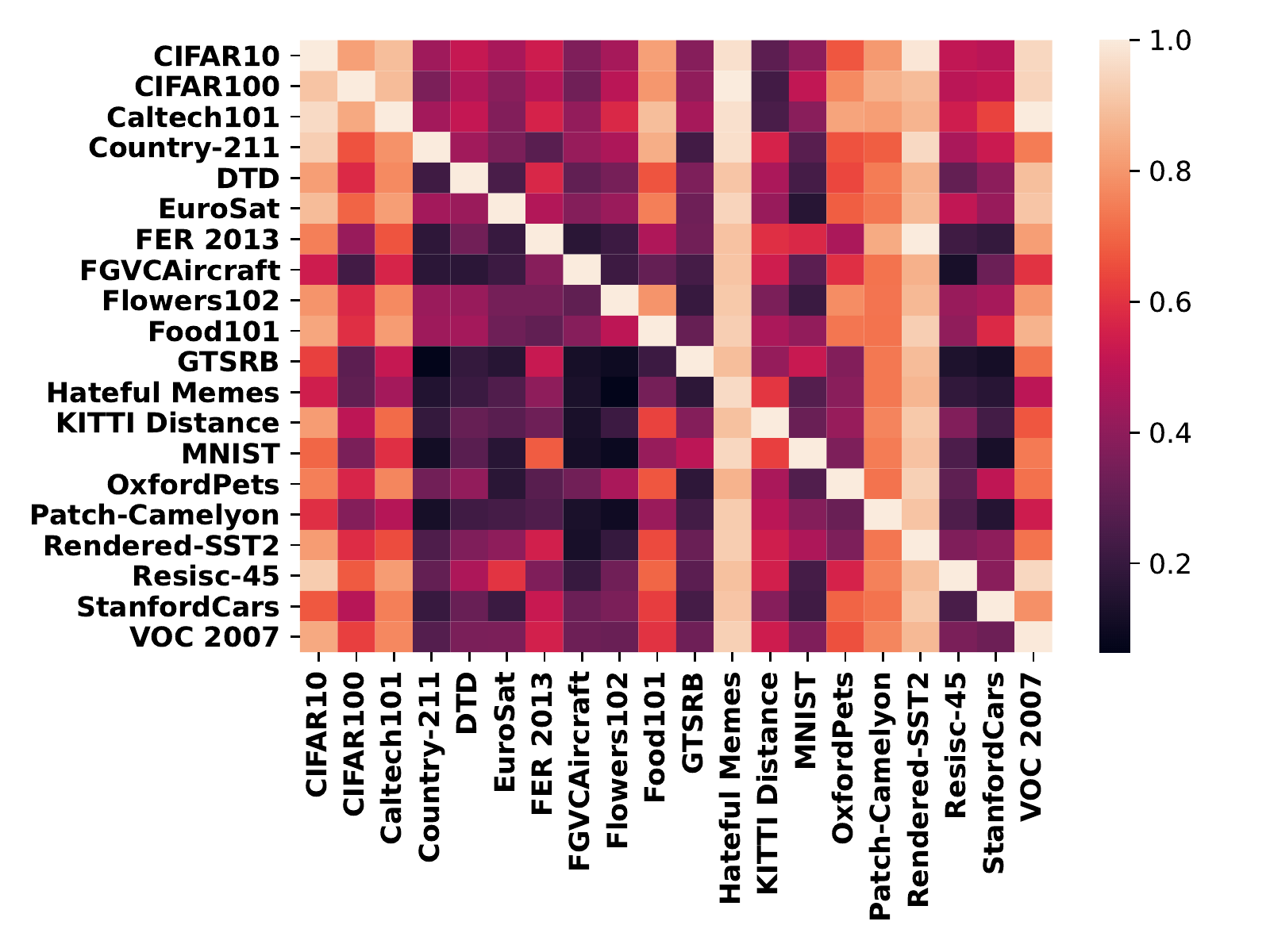}
\caption{\vlpt Transferability.}
\end{subfigure}
\caption{%
A heatmap of our task transferability results. Each cell shows the relative performance on the target task of the transferred prompt from the associated source task (row) to the associated target task (column).}
%\vspace{-2mm}
\label{fig:transfer_heat_map}
\end{figure*}

We first revisit the CLIP~\cite{radford2021learning}, in company with text, visual, and unified prompt tuning approaches for visual recognition in Section.~\ref{subsec:preliminaries}.
We then present technical details of our proposed \ours{} learning in Section.~\ref{subsec:multitask_vl_prompt}.

\subsection{Preliminaries}
\label{subsec:preliminaries}
\noindent \textbf{CLIP} jointly trains an image encoder $\psi$ and a text encoder $\phi$~\cite{radford2021learning}. During pretraining, a image-text pair is encouraged to be mapped to a similar embedding space. CLIP minimizes the symmetric contrastive loss~\cite{chen2020simple,he2020momentum} to predict the positive sample in a batch of image-text combinations:
\begin{equation}
    \label{eq:cos}
    l_{i}^{u\rightarrow v}=\frac{\exp \left(\cos \left(\vu_{i}, \vv_{i}\right) / \tau\right)}{\sum_{j=1}^N \exp \left(\cos \left(\vu_{i}, \vv_{j}\right) / \tau\right)}
\end{equation}
where $\vu = \psi(\vx) \in \mathbb{R}^{d}$ indicating the projection of image $\vx$ to the final hidden space of dimension $d$; $\vv = \phi(\vy) \in \mathbb{R}^{d}$ indicating the projection of text $\vy$; $\cos(\cdot, \cdot)$ denotes the cosine similarity; $\tau$ is a learnable temperature value.
During zero-shot prediction, CLIP is given an image and a set of target classes. Each class is then construct into a fixed prompt ``\texttt{a photo of a [CLASS]}''. The prediction is made by choosing the maximum cosine similarity between the encoded image and the set of prompts.

\noindent \textbf{Text Prompt Tuning.} Since the entire CLIP is usually expensive to finetune, text prompt tuning efficiently leverages tunable text prompts for adapting to downstream tasks. CoOp~\cite{zhou2021coop} proposed a method for adapting CLIP-like vision-language models that replacing a prompt's context words with a learnable vector. Therefore, the text input to encoder is changed to:
\begin{equation} \label{eq:text_prompt}
    \hat{\mP} = [\vp_1, \vp_2, \ldots, \vp_n, \texttt{CLASS}].
\end{equation}
where $\vp_1, \vp_2, \ldots, \vp_n$ is the tunable part and referred as $\mP \in \mathbb{R}^{d \times n}$. $n$ stands for the context prompt length which is adjustable. By modifying the text input, the image and text encoder can be frozen, while only optimizing $\mP$ with each task-specific objective function is sufficient.

\noindent \textbf{Visual Prompt Tuning.} Similar to text prompt tuning, VPT~\cite{jia2022visual} introduces vision prompt tuning. For each $i$-th transformer layer, a tunable part $\mV \in \mathbb{R}^{d \times n}$ is inserted to the original input $[c^{i}, \vq_1, \vq_2, \ldots, \vq_m]$:
\begin{equation} \label{eq:visual_prompt}
    \hat{\mQ}^{i} = [c^{i}, \vv_1, \vv_2, \ldots, \vv_n, \vq_1, \vq_2, \ldots, \vq_m].
\end{equation}
where $c^{i}$ means the classification token \texttt{[CLS]}, and $\vq_1, \vq_2, \ldots, \vq_m$ is patchified image tokens of length $m$. During finetuning, image encoder is frozen and only the visual prompt is optimized.

\noindent \textbf{Unified Prompt Tuning.} Recently, \cite{zang2022unified} proposes Unified Prompt Tuning (\vlpt) approach for adapting VL models. 
\footnote{Due to the recency, \cite{zang2022unified} does not release their model details or code. We therefore  implement our own variant that simply concatenates the CoOp prompt vectors $\mU_T$ and VPT-deep prompt vector $\mU_V$ together as $\mU$, we set the context length of $\mU_T$ and $\mU_V$ the same as 4 unless specify. We use a one-layer one-head Transformer block $\theta$ whose hidden dimension is cut to be 128. Before and after feeding $\mU$ to $\theta$, a linear layer is employed to match the dimensionality. We ablate this design choice in Appendix.} 
Specifically, instead of introducing two sets of isolated modality-specific prompts (\ie, $\mP$ in Eq.~(\ref{eq:text_prompt}) and $\mV$ in Eq.~(\ref{eq:visual_prompt})) for the text and visual encoders, 
\vlpt considers learning a set of vision-language modality-agnostic prompts for tuning VL models. 
\vlpt defines a set of learnable prompts $\mU = [\mU_T, \mU_V] \in \mathbb{R}^{d \times n}$ with length $n$, where $\mU_T  \in \mathbb{R}^{d \times n_T}$, $\mU_V \in \mathbb{R}^{d \times n_V}$ is later employed as textual prompt and visual prompts, respectively. 
A lightweight Transformer layer $\theta$ is used to transform and interact with vision-language prompts $\mU$ before appending the vision-language prompts into the text and visual encoders: 
\begin{equation}
    \begin{aligned}
    \label{eq:gau-gating}
    \mU^{\prime} &= \operatorname{SA}\left( \mU \right) + \mathrm{LN}\left( \mU \right), \\
    \hat{\mU} &= \operatorname{FFN}\left(\mathrm{LN}\left(\mU^{\prime}\right)\right) + \mathrm{LN}\left(\mU^{\prime}\right),
    \end{aligned}
\end{equation}

where the self-attention operator~$\operatorname{SA}$, feed-forward network~$\operatorname{FFN}$ and layer normalization~$\mathrm{LN}$ are applied to obtain the transformed prompts $\hat{\mU}$. The self-attention module in the lightweight Transformer layer allows beneficial interaction between two modalities so as to maximize the complementary effects. 
During downstream training, \vlpt froze both the text and visual encoder ($\phi$ and $\psi$) and only optimizes the vision-language prompts $\mU$ and the lightweight Transformer layer $\theta$. 
In this way, both the dynamic classifiers $\mW$ and visual features $\vz$ in Eq.~(\ref{eq:cos}) are effectively tuned for reliable prediction in the downstream task.

\subsection{Multitask Vision-Language Prompt Tuning}
\label{subsec:multitask_vl_prompt}
Our proposed framework \ours{} mainly consists of two stages as shown in Figure~\ref{fig:approach}, \textit{multitask source prompt initialization} and \textit{multitask target prompt adaptation}. 

\paragraph{Multitask Prompt Initialization.} In this stage, the shareable prompts for all \textit{source tasks} are pretrained jointly through multitask prompt tuning. 
Note that we only use few-shot training set from  \textit{source tasks} to perform this pretrain versus using the entire set in NLP community~\cite{vu2022spot,asai2022attentional}. 

\paragraph{Multitask Prompt Adaptation.} In this stage, we transfer the shareable source prompt to target tasks. 
For single-task target prompt adaptation, we then directly use the learned source prompt to initialize the target prompt and optimize with the regular task loss on each task (\ie, cross-entropy loss). 
For multitask prompt adaptation, we first group relevant tasks together, then perform multitask prompt tuning within the selected groups from the same multitask-initialized source prompt. 
The grouping strategies are further discussed in Section~\ref{subsec:multitask_task_transfer}.

\begin{table*}[t]
    \tabstyle{6pt}
    \caption{\textbf{Comparison of CoOp, VPT, \vlpt, and our MCoOp, MVPT, and M\vlpt in the cross-task generalization setting}. 
    The results strongly justify the \textbf{strong generalizability} of multitask prompt initialization. 
    Specifically, each multitask variant learns shared prompt vectors from 11 \textit{source tasks} before single task adaptation to 12 \textit{target tasks}. 
    The shots number (1, 5, 20) denotes both the number of shots we use for multitask prompt initialization and single task adaptation. 
    For instance, 1 shot means we use 1 shot from each \textit{source task} for multitask prompt initialization and adapt that for 1 shot learning to each \textit{target task}.  
    \textbf{Boldface} text denotes the best performance in that setting.
    Noted that we include the CIFAR-10 in the averaged task table and the CIFAR-10 task performance is included in Appendix.
    }
    \label{tab:crosstask_generalization}
    \begin{subtable}[t]{.38\textwidth} 
    \centering 
    \caption{\textbf{Average over 12 tasks.}}
%   \footnotesize
    \begin{tabular}{l ccccc} 
    \toprule 
    \# shots & 1 & 5 & 20 \\ 
    \midrule 
    CoOp & 50.51$_{\pm1.8}$ & 55.50$_{\pm2.1}$ & 65.87$_{\pm0.5}$ \\ 
    VPT & 57.06$_{\pm1.3}$ & 60.14$_{\pm1.0}$ & 66.98$_{\pm0.7}$ \\ 
    \vlpt & 56.76$_{\pm0.7}$ & 62.16$_{\pm0.8}$ & 67.62$_{\pm0.6}$ \\ 
    \midrule \rowcolor{tabhighlight}
    MCoOp & 55.85$_{\pm1.1}$ & 61.54$_{\pm1.6}$ & 67.60$_{\pm0.5}$  \\ \rowcolor{tabhighlight}
    MVPT & 60.98$_{\pm0.4}$ & \textbf{65.91$_{\pm0.4}$} & 71.73$_{\pm0.3}$  \\ \rowcolor{tabhighlight}
    M\vlpt & \textbf{61.66$_{\pm0.2}$} & 65.77$_{\pm0.4}$ & \textbf{72.15$_{\pm0.4}$}   \\ 
    \bottomrule 
    \end{tabular} 
    \end{subtable} 
    \vspace{1em} 
    \begin{subtable}[t]{.3\textwidth} 
    \centering 
    \caption{CIFAR-100.} 
    \begin{tabular}{l ccccc} 
    \toprule 
    \# shots & 1 & 5 & 20 \\ 
    \midrule 
    CoOp & 64.65 & 70.48 & 72.90 \\ 
    VPT & 70.29 & 73.01 & 77.02 \\ 
    \vlpt & 69.12 & 72.50 & 75.98 \\ 
    \midrule \rowcolor{tabhighlight}
    MCoOp & 63.03 & 71.14 & 72.14   \\ \rowcolor{tabhighlight}
    MVPT & 70.67 & 72.71 & 77.22    \\ \rowcolor{tabhighlight}
    M\vlpt & \textbf{71.17} & \textbf{73.66} & \textbf{77.45}   \\ 
    \bottomrule 
    \end{tabular} 
    \end{subtable} 
    ~ 
    \begin{subtable}[t]{.3\textwidth} 
    \centering 
    \caption{Hateful Memes.} 
    \begin{tabular}{l ccccc} 
    \toprule 
    \# shots & 1 & 5 & 20 \\ 
    \midrule 
    CoOp & 48.40 & 52.60 & 52.40 \\ 
    VPT & 55.40 & 53.20 & 57.20 \\ 
    \vlpt & 51.80 & 54.93 & 56.60 \\ 
    \midrule \rowcolor{tabhighlight}
    MCoOp & 54.00 & 53.80 & 59.40   \\ \rowcolor{tabhighlight}
    MVPT & \textbf{56.20} & \textbf{55.27} & \textbf{57.60}     \\ \rowcolor{tabhighlight}
    M\vlpt & \textbf{56.20} & 55.20 & 56.60     \\ 
    \bottomrule 
    \end{tabular} 
    \end{subtable} 
    ~ 
    \begin{subtable}[t]{.3\textwidth} 
    \centering 
    \caption{MNIST.} 
    \begin{tabular}{l ccccc} 
    \toprule 
    \# shots & 1 & 5 & 20 \\ 
    \midrule 
    CoOp & 49.98 & 78.31 & 91.79 \\ 
    VPT & 71.61 & 74.00 & 88.62 \\ 
    \vlpt & 60.44 & 81.64 & 89.88 \\ 
    \midrule \rowcolor{tabhighlight}
    MCoOp & 65.06 & 78.30 & 94.14   \\ \rowcolor{tabhighlight}
    MVPT & \textbf{82.36} & \textbf{89.57} & \textbf{95.31}     \\ \rowcolor{tabhighlight}
    M\vlpt & 81.29 & 88.48 & 94.54  \\ 
    \bottomrule 
    \end{tabular} 
    \end{subtable} 
    \vspace{1em} 
    \begin{subtable}[t]{.3\textwidth} 
    \centering 
    \caption{Resisc-45.} 
    \begin{tabular}{l ccccc} 
    \toprule 
    \# shots & 1 & 5 & 20 \\ 
    \midrule 
    CoOp & 68.65 & 78.23 & 84.25 \\ 
    VPT & 69.08 & 68.47 & 83.94 \\ 
    \vlpt & 63.68 & 77.25 & 84.05 \\ 
    \midrule \rowcolor{tabhighlight}
    MCoOp & 67.39 & \textbf{79.70} & \textbf{85.12}     \\ \rowcolor{tabhighlight}
    MVPT & \textbf{70.58} & 77.79 & 84.63   \\ \rowcolor{tabhighlight}
    M\vlpt & 70.23 & 77.94 & 85.06  \\ 
    \bottomrule 
    \end{tabular} 
    \end{subtable} 
    ~ 
    \begin{subtable}[t]{.3\textwidth} 
    \centering 
    \caption{Country-211.} 
    \begin{tabular}{l ccccc} 
    \toprule 
    \# shots & 1 & 5 & 20 \\ 
    \midrule 
    CoOp & 12.16 & 21.63 & 22.76 \\ 
    VPT & \textbf{13.76} & 18.26 & 20.71 \\ 
    \vlpt & 13.62 & 21.62 & 21.11 \\ 
    \midrule \rowcolor{tabhighlight}
    MCoOp & 11.75 & \textbf{22.04} & \textbf{23.56}     \\ \rowcolor{tabhighlight}
    MVPT & 11.85 & 17.40 & 19.81    \\ \rowcolor{tabhighlight}
    M\vlpt & 11.37 & 21.33 & 23.53  \\ 
    \bottomrule 
    \end{tabular} 
    \end{subtable} 
    ~ 
    \begin{subtable}[t]{.3\textwidth} 
    \centering 
    \caption{VOC 2007 Classification.} 
    \begin{tabular}{l ccccc} 
    \toprule 
    \# shots & 1 & 5 & 20 \\ 
    \midrule 
    CoOp & 55.78 & 63.70 & 77.43 \\ 
    VPT & 77.54 & 75.91 & 80.59 \\ 
    \vlpt & 79.57 & 76.10 & 78.88 \\ 
    \midrule \rowcolor{tabhighlight}
    MCoOp & 75.84 & 75.46 & 77.60   \\ \rowcolor{tabhighlight}
    MVPT & 78.39 & 79.19 & \textbf{81.67}   \\ \rowcolor{tabhighlight}
    M\vlpt & \textbf{80.18} & \textbf{80.51} & 80.92    \\ 
    \bottomrule 
    \end{tabular} 
    \end{subtable} 
    \vspace{1em} 
    \begin{subtable}[t]{.3\textwidth} 
    \centering 
    \caption{Patch-Camelyon.} 
    \begin{tabular}{l ccccc} 
    \toprule 
    \# shots & 1 & 5 & 20 \\ 
    \midrule 
    CoOp & 59.71 & 51.93 & 59.65 \\ 
    VPT & 56.85 & 57.24 & 57.06 \\ 
    \vlpt & 56.89 & 55.44 & 60.30 \\ 
    \midrule \rowcolor{tabhighlight}
    MCoOp & 52.39 & 56.08 & 69.78   \\ \rowcolor{tabhighlight}
    MVPT & 59.06 & \textbf{66.17} & \textbf{78.10}  \\ \rowcolor{tabhighlight} 
    M\vlpt & \textbf{62.30} & 64.84 & 73.53     \\ 
    \bottomrule 
    \end{tabular} 
    \end{subtable} 
    ~ 
    \begin{subtable}[t]{.3\textwidth} 
    \centering 
    \caption{Rendered-SST2.} 
    \begin{tabular}{l ccccc} 
    \toprule 
    \# shots & 1 & 5 & 20 \\ 
    \midrule 
    CoOp & 55.85 & 54.15 & 54.75 \\ 
    VPT & 57.28 & 54.13 & 57.55 \\ 
    \vlpt & 52.72 & 55.83 & 57.66 \\ 
    \midrule \rowcolor{tabhighlight}
    MCoOp & 56.40 & 56.67 & 57.77   \\ \rowcolor{tabhighlight}
    MVPT & \textbf{59.09} & 58.59 & 59.80   \\ \rowcolor{tabhighlight}
    M\vlpt & 56.07 & \textbf{61.54} & \textbf{61.18}    \\ 
    \bottomrule 
    \end{tabular} 
    \end{subtable} 
    ~ 
    \begin{subtable}[t]{.3\textwidth} 
    \centering 
    \caption{GTSRB.} 
    \begin{tabular}{l ccccc} 
    \toprule 
    \# shots & 1 & 5 & 20 \\ 
    \midrule 
    CoOp & 37.55 & 61.71 & 71.52 \\ 
    VPT & 52.58 & 72.42 & 86.17 \\ 
    \vlpt & \textbf{57.67} & 70.72 & 85.34 \\ 
    \midrule \rowcolor{tabhighlight}
    MCoOp & 37.89 & 59.31 & 72.09   \\ \rowcolor{tabhighlight}
    MVPT & 50.56 & \textbf{75.03} & \textbf{89.75}  \\ \rowcolor{tabhighlight} 
    M\vlpt & 51.79 & 69.22 & 85.30  \\ 
    \bottomrule 
    \end{tabular} 
    \end{subtable} 
    \vspace{1em} 
    \begin{subtable}[t]{.3\textwidth} 
    \centering 
    \caption{FER 2013.} 
    \begin{tabular}{l ccccc} 
    \toprule 
    \# shots & 1 & 5 & 20 \\ 
    \midrule 
    CoOp & 29.34 & 28.25 & 50.71 \\ 
    VPT & 49.76 & 47.48 & 56.39 \\ 
    \vlpt & 49.76 & 47.85 & 56.77 \\ 
    \midrule \rowcolor{tabhighlight}
    MCoOp & 52.49 & 47.76 & 50.24   \\ \rowcolor{tabhighlight}
    MVPT & 51.43 & 50.85 & 57.12    \\ \rowcolor{tabhighlight}
    M\vlpt & \textbf{55.95} & \textbf{51.27} & \textbf{60.07}   \\ 
    \bottomrule 
    \end{tabular} 
    \end{subtable} 
    ~ 
    \begin{subtable}[t]{.3\textwidth} 
    \centering 
    \caption{KITTI Distance.} 
    \begin{tabular}{l ccccc} 
    \toprule 
    \# shots & 1 & 5 & 20 \\ 
    \midrule 
    CoOp & 34.60 & 21.38 & 60.90 \\ 
    VPT & 23.77 & 40.79 & 47.68 \\ 
    \vlpt & 37.41 & 42.57 & 53.54 \\ 
    \midrule  \rowcolor{tabhighlight}
    MCoOp & 45.01 & 46.69 & 57.38   \\ \rowcolor{tabhighlight} 
    MVPT & \textbf{52.60} & \textbf{58.46} & 67.65  \\ \rowcolor{tabhighlight}
    M\vlpt & 50.77 & 53.73 & \textbf{73.98}     \\ 
    \bottomrule 
    \end{tabular} 
    \end{subtable} 

\end{table*}

\begin{table*}[h]
    \tabstyle{3pt}
    \caption{\textbf{Comparison of prompt learning methods on the few-shot \elevater }. 
    The number of shots is set to be 20 in each case, except for zero-shot CLIP. 
    The results suggest the significant generalizability of multitask prompt initialization. 
    $^\dagger$ denotes we obtain the zero-shot CLIP results from \elevater~\cite{li2022elevater} 
    ``Source'' denotes the prompt initialization source, where ``-'' stands for random initialization, 
    and ``M'' stands for using all 20 \elevater tasks for prompt initialization. 
    ``Adaptation'' denotes the \textit{target task} prompt adaptation method, where ``S'' stands for single \textit{target task} prompt adaptation that each t\textit{target task} will be adapted independently, and `M`'' stands for multitask prompt adaptation that certain tasks (selected based on results in Section~\ref{subsec:multitask_task_transfer}) will be learned together. 
    Clearly, \ours{} demonstrates better \textbf{transferability} than single \textit{target task} prompt adaptation counterparts. 
    $\Delta$ denotes the best M-variant's gain over the best CoOp/VPT/\vlpt baseline methods. 
    }
    \label{tab:fewshot_elevater}
    \begin{adjustbox}{max width=\textwidth}
    \begin{tabular}{l c cccccccccccccccccccccc}
    \toprule
    &  &  & \multicolumn{21}{c}{Target} \\ 
     \cmidrule(lr){4-23}
    & \rotbox{Source} & \rotbox{Adaptation} & \rotbox{Caltech101} & \rotbox{CIFAR10} & \rotbox{CIFAR100} & \rotbox{Country-211} & \rotbox{DTD} & \rotbox{EuroSat} & \rotbox{FER 2013} & \rotbox{FGVCAircraft} & \rotbox{Flowers102}& \rotbox{Food101} & \rotbox{GTSRB} & \rotbox{Hateful Memes} & \rotbox{KITTI Distance} & \rotbox{MNIST} & \rotbox{OxfordPets} & \rotbox{Patch-Camelyon} & \rotbox{Rendered-SST2} & \rotbox{Resisc-45} & \rotbox{StanfordCars} & \rotbox{VOC 2007} & \rotbox{\emph{Average}}  \\
    \midrule
    CLIP$^\dagger$ & - & - & 88.9 & 90.8 & 68.2 & 22.8 & 44.8 & 54.7 & 48.5 & 24.3 & 88.7 & 43.5 & 58.1 & 27.0 & 52.0 & 69.4 & 89.0 & 54.0 & 60.9 & 65.6 & 64.8 & 83.7 & 60.0 \\
    \midrule
        CoOp & - & S & 91.44 & 91.30 & 73.01 & 22.83 & 69.82 & 80.19 & 54.46 & 42.01 & 93.31 & 89.47 & 73.87 & 52.40 & 56.87 & 91.44 & 90.69 & 62.79 & 59.55 & 83.83 & 79.52 & 74.61 & 71.67$_{\pm0.2}$ \\
    VPT & - & S & 92.84 & 91.39 & 75.98 & 21.11 & 68.56 & 87.37 & 56.77 & 42.12 & 89.22 & 89.04 & 85.34 & 56.60 & 53.54 & 89.88 & 90.71 & 60.30 & 57.66 & 84.05 & 74.95 & 78.88 & 72.32$_{\pm0.6}$\\
    \vlpt & - & S & 92.58 & 92.05 & 76.61 & 23.37 & 67.68 & 88.98 & 56.87 & 42.46 & 89.59 & 89.64 & 82.72 & 56.87 & 47.87 & 89.11 & 91.24 & 60.41 & 59.03 & 83.32 & 76.40 & 81.20 & 72.40$_{\pm0.3}$\\
    \midrule \rowcolor{tabhighlight}
        MCoOp & - & M & 91.53 & 91.67 & 73.01 & 23.12 & 69.82 & 81.69 & 54.46 & 42.01 & 93.44 & 89.47 & 74.38 & 58.40 & 56.87 & 91.44 & 90.69 & 64.91 & 61.63 & 84.03 & \textbf{79.52} & 78.45 & 72.53$_{\pm0.6}$\\ \rowcolor{tabhighlight}
    MVPT & - & M & 92.84 & 93.54 & 76.39 & 21.42 & 68.56 & 89.15 & 56.77 & 42.12 & 89.22 & 89.04 & 85.34 & 58.20 & 53.54 & 89.88 & 91.01 & 66.53 & 58.14 & 84.05 & 74.95 & 80.69 & 73.07$_{\pm0.9}$ \\ \rowcolor{tabhighlight}
    M\vlpt & - & M & 92.58 & 93.38 & 76.61 & 23.37 & 67.68 & 88.98 & 56.94 & 42.46 & 89.59 & \textbf{89.64} & 82.72 & 58.13 & 55.41 & 89.91 & \textbf{91.24} & 63.36 & 61.34 & 83.32 & 76.40 & 81.20 & 73.21$_{\pm0.7}$\\
    \midrule \rowcolor{tabhighlight}
    MCoOp & M & M & 92.09 & 91.59 & 72.63 & \textbf{23.52} & \textbf{70.41} & 81.70 & 54.85 & 42.34 & \textbf{93.61} & 89.14 & 72.74 & 58.40 & 47.73 & 90.21 & 89.61 & 68.92 & \textbf{64.89} & \textbf{84.39} & 79.43 & 79.55 & 72.39$_{\pm0.5}$\\ \rowcolor{tabhighlight}
    MVPT & M & M & \textbf{93.46} & 93.72 & \textbf{77.38} & 20.79 & 69.43 & \textbf{92.23} & \textbf{57.07} & \textbf{42.57} & 88.80 & 87.78 & \textbf{89.62} & 55.53 & \textbf{62.07} & \textbf{93.08} & 91.04 & 69.69 & 57.50 & 84.35 & 74.20 & \textbf{82.21} & \textbf{74.13}$_{\pm0.3}$ \\ \rowcolor{tabhighlight}
    M\vlpt & M & M & 92.19 & \textbf{93.75} & 75.39 & 23.45 & 65.99 & 90.17 & 56.06 & 41.19 & 89.34 & 89.38 & 81.66 & \textbf{59.00} & 57.20 & 91.38 & 90.30 & \textbf{69.74} & 62.29 & 83.40 & 76.66 & 79.29 & 73.39$_{\pm0.6}$ \\
    \midrule
    $\Delta$ & & & \hgreen{+0.62} & \hgreen{+1.70} & \hgreen{+0.77} & \hgreen{+0.15} & \hgreen{+0.59} & 
    \hgreen{+3.25} & \hgreen{+0.20} & \hgreen{+0.11} & \hgreen{+0.30} & \hblue{+0.00} & \hgreen{+4.28} & \hgreen{+2.13} & \hgreen{+5.20} & \hgreen{+1.64} & \hblue{+0.00} & \hgreen{+6.95} & \hgreen{+5.34} & \hgreen{+0.34} & \hblue{+0.00} & \hgreen{+1.01} &\hgreen{+1.73} \\
    \bottomrule
    \end{tabular}
    \end{adjustbox}
\end{table*}
\section{Experiments}
\label{sec:experiment}

Our approach is mainly evaluated in the following three problem settings: 
1) cross-task generalization (Section~\ref{subsec:multitask_prompt_init}) that measures the efficacy of multitask prompt initialization; 
2) few-shot \elevater (Section~\ref{subsec:multitask_prompt_adapt}) that shows the effectiveness of multitask prompt adaption; and, 
3) zero-shot task transferability (Section~\ref{subsec:multitask_task_transfer}) that is based on the 20 vision tasks in \elevater. 
All models used in our experiments are based on the open-source pretrained CLIP model~\cite{radford2021learning}.\footnote{\url{https://github.com/openai/CLIP}.} Before discussing the results, we provide the details of the experimental setup below.

\paragraph{Datasets}
For the domain generalization setting, we use the 11 image recognition tasks from ~\cite{zhou2021coop} as \textit{source tasks}.
We use the non-overlapped 12 image recognition tasks in \elevater~\cite{li2022elevater} as \textit{target tasks}, covering a diverse set of recognition tasks. 
Specifically, the \textit{source tasks} include ImageNet~\cite{deng2009imagenet} and Caltech101~\cite{fei2004learning} for generic objects classification; OxfordPets~\cite{parkhi2012cats}, StanfordCars~\cite{krause20133d}, Flowers102~\cite{nilsback2008automated}, Food101~\cite{bossard2014food} and FGVCAircraft~\cite{maji2013fine} for fine-grained classification; SUN397~\cite{xiao2010sun} for scene recognition; UCF101~\cite{soomro2012ucf101} for action recognition; DTD~\cite{cimpoi2014describing} for texture classification; and, EuroSAT~\cite{helber2019eurosat} for satellite imagery recognition. 
The \elevater benchmarks originally cover 20 image classification tasks which includes the 8 overlapped tasks as Caltech101, OxfordPets, StanfordCars, Flowers102, Food101, FGVCAircraft, DTD and EuroSAT, and the rest 12 non-overlapped tasks as Hateful Memes~\cite{kiela2020hateful}, PatchCamelyon~\cite{veeling2018rotation}, Rendered-SST2~\cite{radford2021learning}, KITTI Distance~\cite{fritsch2013new}, FER 2013~\cite{fer2013}, CIFAR-10/100~\cite{krizhevsky2009learning}, VOC 2007 Classification~\cite{everingham2010pascal}, Country-211~\cite{radford2021learning},  MNIST~\cite{deng2012mnist}, GTSRB~\cite{stallkamp2011german}, and Resisc-45~\cite{cheng2017remote}. 
We use the 12 non-overlapped classification tasks as \textit{target tasks} in the first setting and all 20 tasks in the second few-shot learning setting and the third zero-shot transferability setting.  
Following \elevater~\cite{li2022elevater}, we randomly sample for each dataset a few-shot training set while using the original test set for testing. 
We evaluate across 0, 1, 5 and 20 shot numbers as in \elevater benchmarks. For learning-based models, the results are averaged over three runs.

\paragraph{Baselines}
We compare our approach against the following methods:
(\romannum{1}) \textbf{Zero-shot CLIP}. This baseline uses does not involve any prompt-learning strategies as mentioned in Section~\ref{subsec:preliminaries}.
(\romannum{2}) \textbf{Single Task Prompt Tuning} methods, including CoOp~\cite{zhou2021coop}, VPT~\cite{jia2022visual}, \vlpt~\cite{zang2022unified} for vision, language, and vision-language prompt tuning method, respectively.

\paragraph{Training Details}
Our implementation is based on CoOp.\footnote{\url{https://github.com/KaiyangZhou/CoOp}.} 
Throughout the experiments, we use CLIP as our vision-language model (\ie, ViT-B/16 for all the experiments except for the scaling ablation). 
Following CoOp~\cite{zhou2021coop} and VPT~\cite{jia2022visual}, we use a context length of 16 for both CoOp and VPT throughout the study. 
We empirically find a shorter context length of 4 leads to better performance for \vlpt, and we use 4 contexts for \vlpt only. (This design choice is discussed in more detail in the Appendix). 
The resulting prompt vectors of CoOp/MCoOp, VPT/MVPT, \vlpt/M\vlpt  account for 0.01\%, 0.11\%, 0.45\% total parameters of the ViT-B/16 (124M parameters) model. 
All the prompt vectors for CoOp, VPT or \vlpt are randomly initialized without  using the pretrained word embeddings of ``a photo of a'' for initialization in \cite{zhou2021coop} for a fair comparison. 
All the methods are trained with a batch size of 32 for 200 epochs following \cite{zhou2021coop}. 
All the image input size is set to 224$\times$224. 
We use Adam optimizer and cosine learning rate schedule. All the learning rate is set as 2e-3, and the warmup period is set as 1 epoch following \cite{zhou2021coop}. 
All the few-shot experiments are averaged with 3 runs. 
For each experiment, we select the best prompt checkpoint using the validation set that consists 20\% splits from the few-shot sampled training set.

\subsection{Cross-task Generalization}
\label{subsec:multitask_prompt_init}
% \paragraph{xxx helps Domain Generalization (ELEVATER-CoOp)}
We examine the efficacy of the proposed multitask prompt initialization in \ours{} via cross-task generalization. 
Specifically, we use all the 11 tasks in~\cite{zhou2021coop} as \textit{source tasks} and the non-overlapped 12 tasks in \elevater as \textit{target tasks}. 
We perform multitask learning on all the \textit{source tasks} to learn the shared prompt vectors. 
The resulting shared prompt vectors will be used as the prompt initialization for single-task adaptation on each \textit{target task}. 
We evaluate across 1, 5, 20 shots as suggested in the \elevater~\cite{li2022elevater} benchmark. 
The shot number is adopted for both multitask prompt initialization and single \textit{target task} adaptation, respectively. 
It means that for 1 shot, we will sample 1 instance for each image class of all the \textit{source tasks} for multitask prompt initialization and then adapt the learned prompt initialization to 1-shot learning for each \textit{target task}. 
The baseline prompt learning method CoOp, VPT and \vlpt are using random initialized prompt as in~\cite{zhou2021coop,jia2022visual} for single \textit{target task} adaptation. 
The results are summarized in Table~\ref{tab:crosstask_generalization}, showing that multitask prompt initialization variants MCoOp, MVPT and M\vlpt mostly outperform the baseline prompt learning counterparts by a significant margin. (averaged over 3 runs).  
The improvement is also consistent across different numbers of shots. 
It is also interesting that the most effective task of multitask prompt initialization differs for each prompt learning method. 
Specifically, MCoOp benefits the task where the class names are  distinct the most like Resisc-45, while MVPT/M\vlpt favors the task where the images are more separable like VOC 2007 Classification. 
We further analyze this different preference in Section~\ref{subsec:multitask_task_transfer}.  
Nevertheless, we note that multitask prompt initialization does not always guarantee performance improvement when the number of \textit{source task} shots is extremely small as 1 and the \textit{target task} needs extreme fine-grained or specialized classification like 211-way classification in Country-211.

\subsection{Few-shot \elevater}
\label{subsec:multitask_prompt_adapt}
% \paragraph{xxx helps Domain Transfer in ELEVATER}
We measure the effectiveness of the proposed multitask prompt adaptation in \ours{} on all 20 few-shot \elevater tasks.  
We set the number of shots as 20 in each setting. 
Specifically, versus adapting the learned prompt initialization to each \textit{target task} independently (single-task prompt adaptation), we group several \textit{target tasks} as in Figure~\ref{fig:approach} and perform multitask learning in each group to learn shared prompt vectors during prompt adaptation. 
We determine which tasks should be grouped for each prompt learning method based on the transferability map shown in Figure~\ref{fig:transfer_heat_map}, which is discussed in more details in Section~\ref{subsec:multitask_task_transfer}. 
The detailed results are shown in Table~\ref{tab:fewshot_elevater}. 
It clearly demonstrates that multitask prompt adaptation variants exhibit better transferability than single \textit{target task} prompt adaptation counterparts.  
Comparing single-task prompt adaption and multitask prompt adaption, multitask adaption boosts the averaged performance on CoOp, VPT, \vlpt by 0.86\%, 0.75\% and 0.81\%, respectively. 
Using 20 \elevater tasks as \textit{source tasks} can further improve the results for MVPT and M\vlpt. For MCoOp, multitask prompt initialization may make the class name distribution less separable for the task has distinct categories like KITTI Distance, which effaces the improvement on other tasks. 
The resulting MVPT achieves \textbf{74.13}\% the new state-of-the-art on 20 shot \elevater benchmark for ViT-B/16 model 
% with input image size as 224$\times$224 
comparing to 64.41\% in \cite{liang2022supmae}. 
% compared to the best \textbf{70.72}\% average performance for ViT-B/16 CLIP model from \elevater
% leaderboard\footnote{\url{https://eval.ai/web/challenges/challenge-page/1832/leaderboard}}. 
We also observe that there exist tasks that are not improved using multitask prompt adaptation. 
We attribute that to 
some tasks like FGVCAircraft with distant and specialized categories may not be able to leverage useful cross-task knowledge from other \elevater tasks during prompt adaptation. 

\subsection{Task Transferability}
\label{subsec:multitask_task_transfer}
It is a long debatable history in vision that ``\textit{which tasks should be learned together?}'' since the introduction of multitask learning~\cite{thrun1995learning,caruana1997multitask,zhang2021survey,standley2020tasks,raffel2020exploring}. 
To understand the cross-knowledge in vision-language prompt tuning further, we conduct a large-scale study on task transferability with 20 \elevater tasks in 400 combinations for each prompt tuning method. 
We use checkpoints from each task in \elevater after 20-shot learning 
on 
% 3 different seeds as 
the \textit{source}. 
Then, we perform zero-shot adaptation to the rest of the tasks. 
We normalized the scores to [0, 1] by dividing the transfer performance with the best one on that task and presented the results in Figure~\ref{fig:transfer_heat_map}. 
It shows that the tasks share similar classnames (CIFAR-10 and Caltech 101) have better transferability using CoOp and similar image space (CIFAR-10 and Flowers102) leads to better transferability with VPT/\vlpt. 
To select groups for multitask adaptation, we select the top 1 and 2 transferability with respect to each \emph{target task}. 
We jointly train such group of 2 and 3 tasks 
and select the best checkpoint based on the the validation performance for each task, respectively. 
We also report the performance with different grouping strategies for multitask prompt adaptation on 20-shot \elevater in Table~\ref{tab:prompt_adapt_strategy}, where Best M stands for using the aforementioned grouping strategy and Worst M stands for grouping the most dissimilar tasks from the transferability map.\footnote{We provide detailed task group information in Appendix.}
It directly suggests that the transferability map could serve as a principal way to group the relevant tasks and failing to do that leads to significant performance degradation.

\begin{table}[t]
    \tabstyle{3pt}
    \caption{Ablation of prompt adaptation strategies for \ours{}.
    }
    \footnotesize
    \label{tab:prompt_adapt_strategy}
    \begin{tabular}{l c c c}
    \toprule
    \multirow{2}{*}{Model} &  \multirow{2}{*}{Source} & \multirow{2}{*}{Adaptation} & Averaged \\
    & & & \elevater \\
    \midrule
    MCoOp & M & S & 70.93$_{\pm0.3}$\\
    MVPT  & M & S & 73.16$_{\pm0.3}$\\
    M\vlpt & M & S & 72.25$_{\pm0.5}$ \\
    \midrule
    MCoOp & M & Best M & \textbf{72.39$_{\pm0.5}$}\\
    MVPT & M &  Best M & \textbf{74.13$_{\pm0.3}$}\\
    M\vlpt & M & Best M & \textbf{73.39$_{\pm0.6}$}\\
    \midrule
    MCoOp & M & Worst M & 70.13$_{\pm0.7}$ \\
    MVPT & M &  Worst M & 71.81$_{\pm0.2}$\\
    M\vlpt & M & Worst M & 69.94$_{\pm1.0}$ \\
    \bottomrule
    \end{tabular}
\end{table}
\section{Discussion}
\label{sec:ablation}
\paragraph{Source Tasks}

There is rich literature~\cite{he2016deep,he2020momentum,chen2020simple,he2022masked} to use ImageNet1K to pretrain vision backbones for various downstream vision tasks (object detection~\cite{girshick2015fast}, semantic segmentation~\cite{he2017mask}).   
In Table~\ref{tab:source_task_impact}, we study the impact of \textit{source tasks} using ImageNet1K, 10 \textit{source tasks} in \cite{zhou2021coop} excluding ImageNet1K and 11 \textit{source tasks} in \cite{zhou2021coop} including ImageNet1K using the 20-shot cross-task generalization setting. 
It shows that ImageNet1K servers as a strong \textit{source task} for prompt initialization while performing multitask prompt initialization from the diverse 10 \textit{source tasks} leads to noticeable improvement especially for MVPT/M\vlpt. 
Besides, combing ImageNet1K with the 10 \textit{source tasks} gives the best performance, which may suggest the potential to scale our \ours{} to more diverse set as \textit{source tasks} like even more than thousands of tasks~\cite{wang2022benchmarking,chung2022scaling} in NLP community. 

\begin{table}[t]
    \tabstyle{3pt}
    \footnotesize
    \caption{Ablation of source tasks for MCoOp, MVPT and \ours{}. 
    }
    \label{tab:source_task_impact}
    \begin{tabular}{l c c c}
    \toprule
    \multirow{2}{*}{Model} &  \multirow{2}{*}{Source} & \multirow{2}{*}{Adaptation} & Averaged \\
    & & & 12 \textit{target tasks} \\
    \midrule
    CoOp & ImageNet1K & S & 66.36$_{\pm0.5}$ \\
    VPT  & ImageNet1K & S & 68.80$_{\pm0.9}$ \\
    \vlpt & ImageNet1K & S & 67.45$_{\pm0.7}$ \\
    \midrule
    MCoOp & \textit{10 source tasks} & S & 66.51$_{\pm0.5}$  \\
    MVPT & \textit{10 source tasks} &  S & 70.31$_{\pm1.1}$ \\
    M\vlpt & \textit{10 source tasks} & S & 70.08$_{\pm0.9}$ \\
    \midrule
    MCoOp & \textit{11 source tasks} & S & \textbf{67.60}$_{\pm0.5}$  \\
    MVPT & \textit{11 source tasks} &  S & \textbf{71.73}$_{\pm0.6}$ \\
    M\vlpt & \textit{11 source tasks} & S & \textbf{72.15}$_{\pm0.7}$ \\
    \bottomrule
    \end{tabular}
\end{table}

\paragraph{Scaling} 
We conduct scaling experiments 
in Figure~\ref{fig:mvlpt_scaling} to analyze how \ours{} performs with increasing pretrained model sizes.
It is based on 20-shot cross-task generalization setting except for 0-shot CLIP, 
These results show that our MVPT, MCoOp, M\vlpt is 
not only able to achieve the same parameter efficiency but also 
effective across model scales ranging from ViT-B/32 to ViT-L/14. 
ViT-B/32 (2.59 GFLOPs, 125M parameters) to ViT-B/16 (11.27 GFLOPs) and ViT-L/14 (51.90 GFLOPs, 390M parameters).

\begin{figure}[t]
\centering
% \begin{subfigure}{.23\textwidth}\includegraphics[width=\textwidth]{figs/ablate_flops.pdf}
% \caption{Scaling experiments.}
% \end{subfigure}
\begin{subfigure}{.3\textwidth}\includegraphics[width=\textwidth]{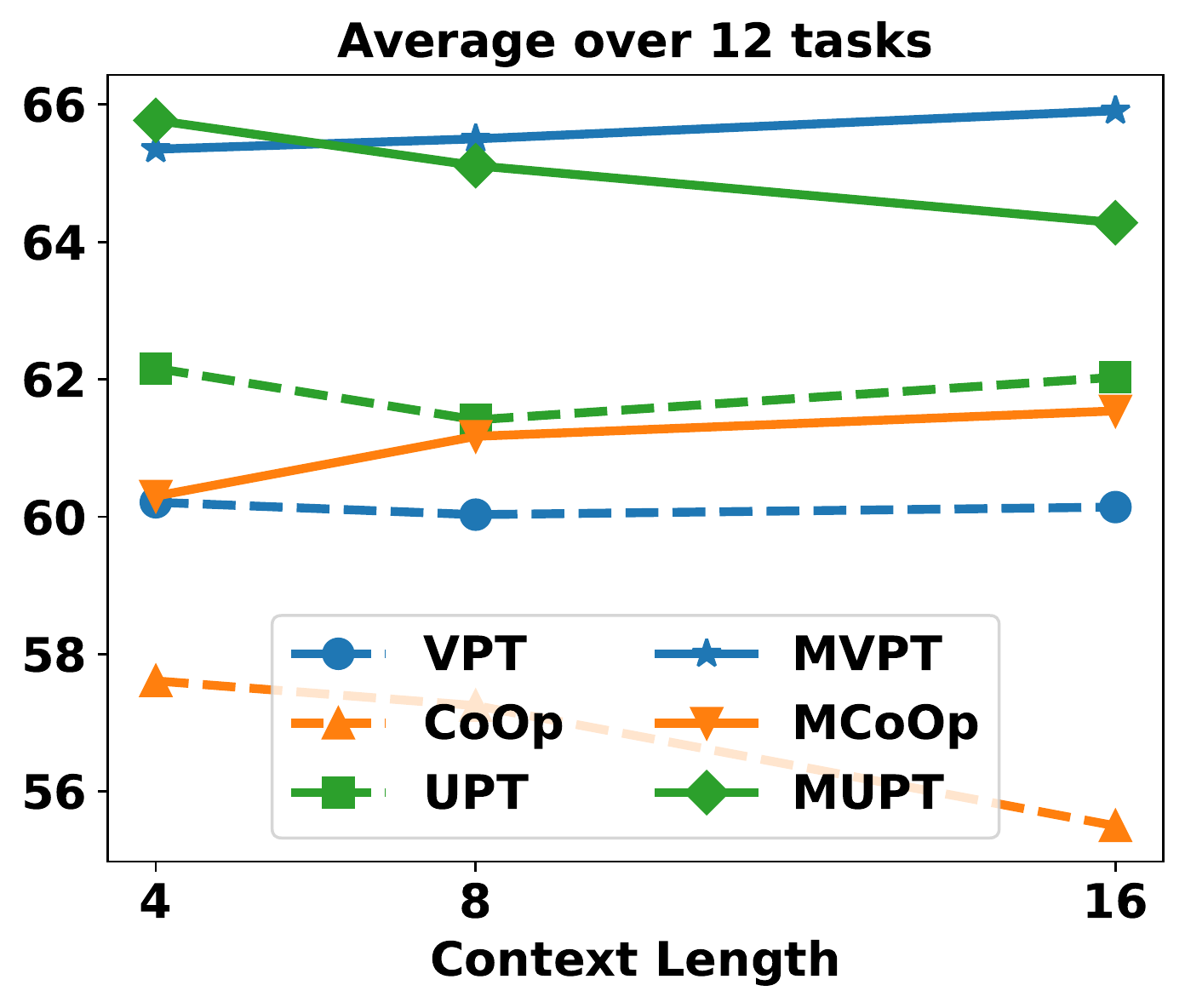}
% \caption{Ablation on context length.}
\end{subfigure}
\caption{Ablation on context length.} %\sheng{@sheng}
\label{fig:ablate_scale_ctxlen}
\vspace{-2mm}
\end{figure}

\paragraph{Context Length} 
The ablation study on context length is also carried out in the cross-task generalization setting. 
Following ~\cite{zhou2021coop}, we study 4, 8 and 16 context tokens. 
% For fair comparison, 
We use random initialization for all context tokens. In Figure~\ref{fig:ablate_scale_ctxlen}, we see consistent improvement for longer context length of MVPT, MCoOp and marginal performance difference for VPT, CoOp, UPT. 
For MUPT, we observe longer context length turns out to hurt the performance, which we assume could be potentially attribute to the context length discussion in CoOp~\cite{zhou2021coop}.  

\paragraph{Limitations}
As discussed in the Section~\ref{sec:method}, the improvements of \emph{multitask prompt initialization} accompanies the cost of extra compute for multitask prompt tuning on \emph{source tasks}.
Even though the procedure is conducted once like pretraining then the learned prompt can be reused as initialization for various target tasks. 
It sums up to $\frac{N_\text{source}}{N_\text{target}}$ more compute ($N_\text{source}$, $N_\text{target}$ stands for number of \emph{source tasks}, \emph{target tasks}, respectively). 
The extra compute caused by \emph{multitask prompt adaptation} is marginal except for evaluating the zero-shot task transferability for task grouping guidance. 

\section{Conclusion}
\label{sec:conclusion}
In this paper, we propose multitask vision-language prompt tuning (\ours{}).
We demonstrate that \ours{} exhibits strong generalizability and few-shot learning performance compared to baseline prompt learning methods (CoOp, VPT, and \vlpt). 
The most performant \ours{} sets the new state-of-the-art performance on the \elevater benchmark. 
We also study task transferability across 20 vision tasks and provide a guideline for multitask prompt learning. 
We show that multitask vision-language prompt tuning leverages the cross-task knowledge and helps the individual task performance on \elevater benchmarks.  
We hope our study will inspire future research on large-scale multitask learning in the vision-language domain and how to adapt to various downstream tasks more effectively. 

\paragraph{Acknowledgements} The authors gratefully acknowledge Chunyuan Li for the early insightful discussions on the use of multi-task learning for vision-language tasks.  
SS and KK are supported by Samsung SAIT, Intel corporation, Intel VLAB team, Intel One-API center of excellence, as well as funding through BDD and BAIR. 
The work of SS and TD was supported in part by DoD including DARPA’s LwLL, PTG programs, as well as BAIR’s industrial alliance programs.

% \clearpage
% \newpage
{\small
\bibliography{egbib}
\bibliographystyle{plain}
}
% \clearpage
% \newpage
\section{Appendix}
\subsection{Additional Results}

\begin{figure*}[t]
\centering
\begin{subfigure}{.33\textwidth}\includegraphics[width=\textwidth]{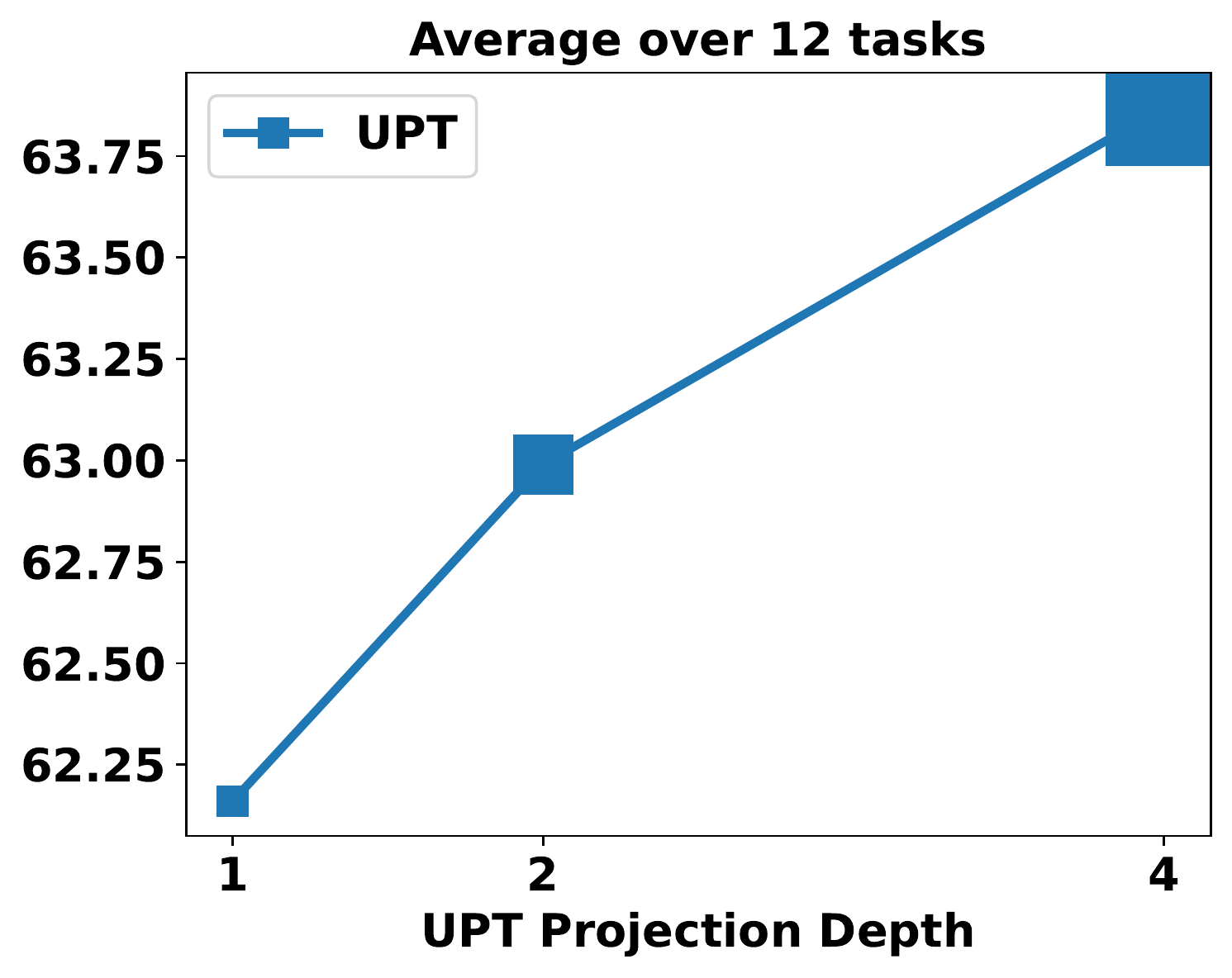}
\caption{Ablation on the UPT Projection Depth.}
\end{subfigure}
\begin{subfigure}{.33\textwidth}\includegraphics[width=\textwidth]{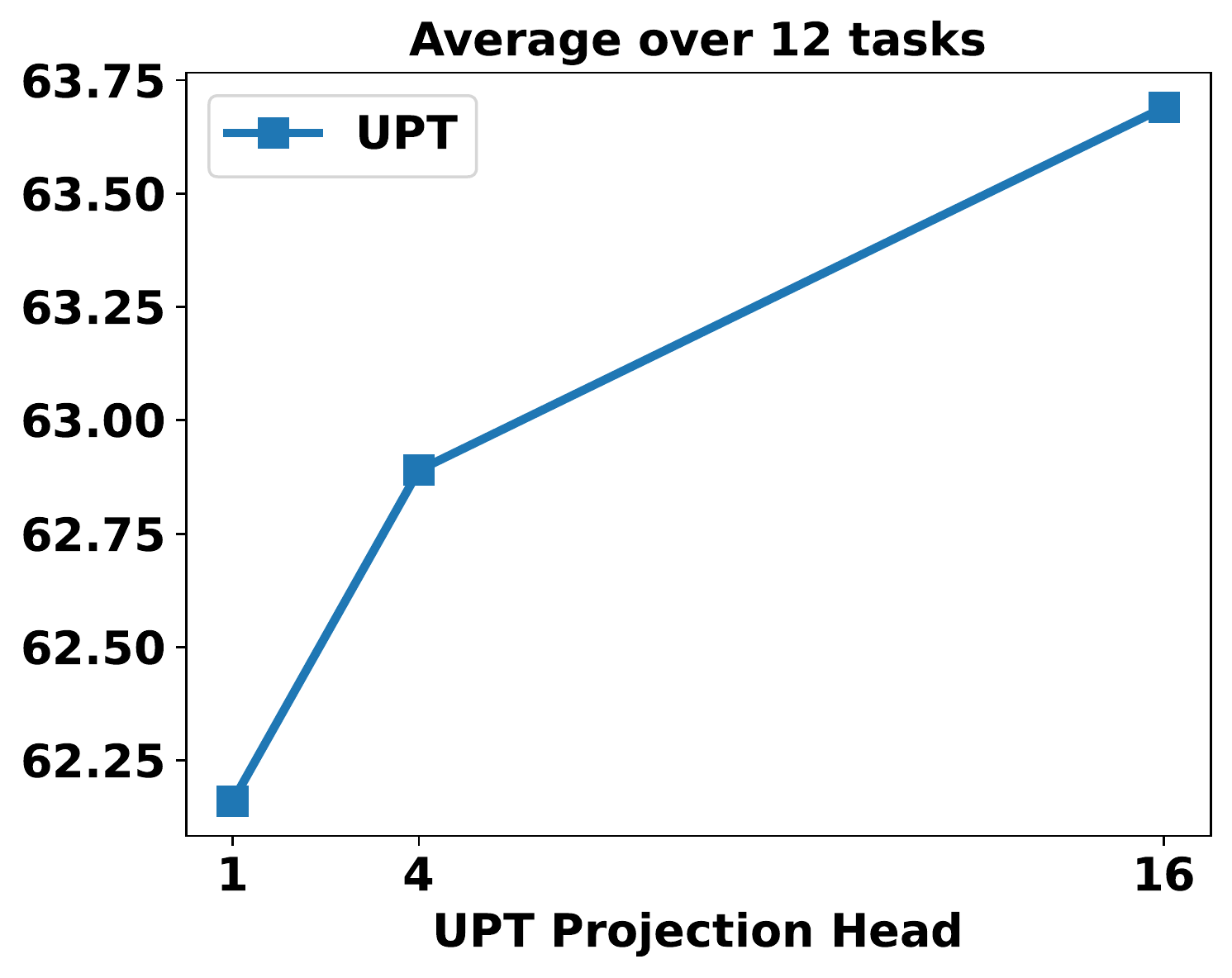}
\caption{Ablation on number of heads.}
\end{subfigure}
\begin{subfigure}{.31\textwidth}\includegraphics[width=\textwidth]{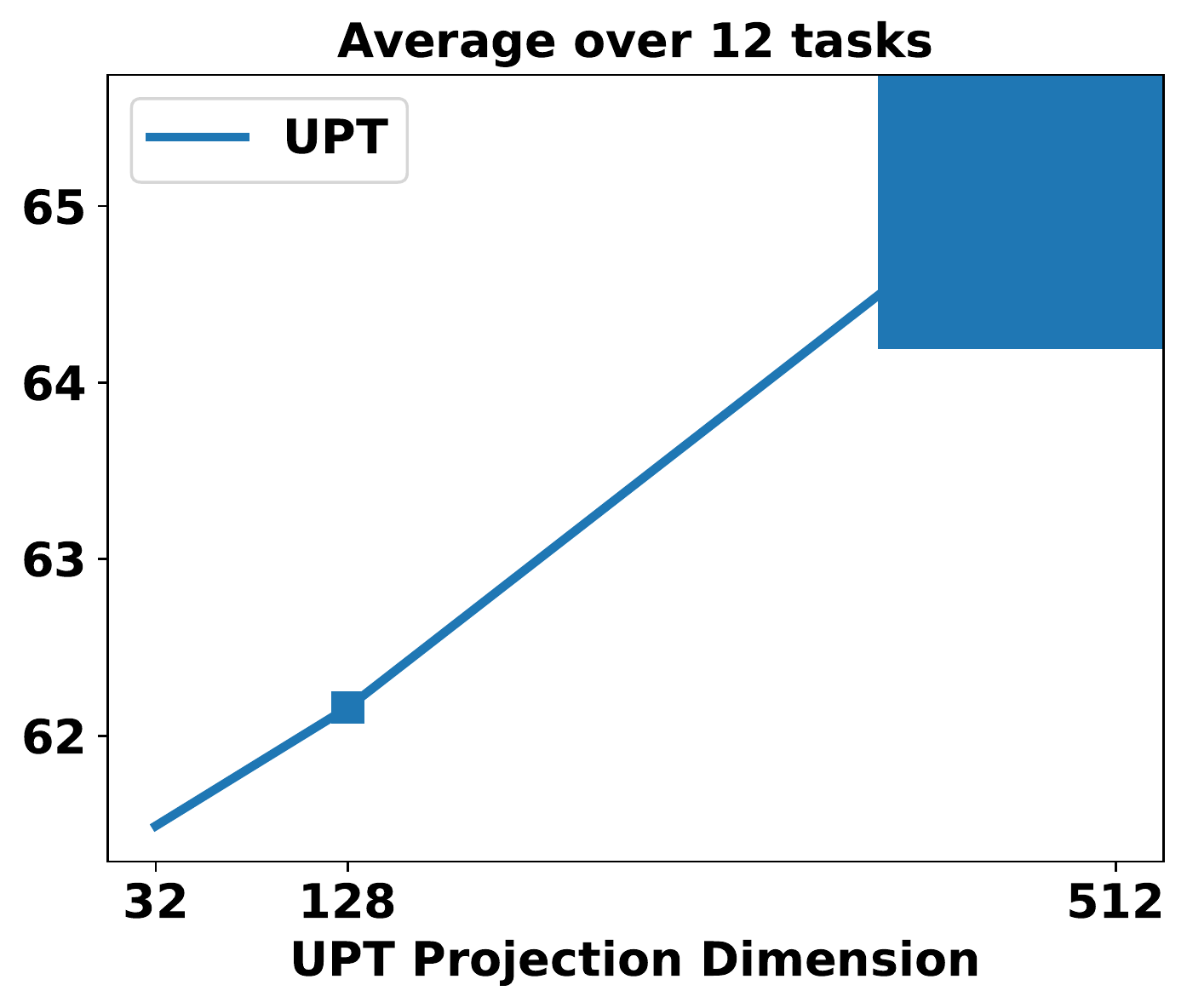}
\caption{Ablation on the UPT Projection Dimension.}
\end{subfigure}
\caption{Ablation on the hyper-parameter of UPT. Note that the size of each point stands for the relative additional parameter size.  }
% \protect\footnotemark} %\sheng{@sheng}
\label{fig:ablate_upt_appendix}
% \vspace{-2mm}
\end{figure*}

\paragraph{Ablation on \vlpt}

As mentioned in the main text, due to the recency, \cite{zang2022unified} does not release their model details or code. We therefore  implement our own variant that simply concatenates the CoOp prompt vectors $\mU_T$ and VPT-deep prompt vector $\mU_V$ together as $\mU$, we set the context length of $\mU_T$ and $\mU_V$ the same as 4 unless specify. We use a one-layer one-head Transformer block $\theta$ whose hidden dimension is cut to be 128. Before and after feeding $\mU$ to $\theta$, a linear layer is employed to match the dimensionality. 
We ablate the design choice on number of heads, number of layer, and dimensionality, respectively in Figure \ref{fig:ablate_upt_appendix}. 
The size of the each point stands for the relative additional parameter size included in this setting. 

% We ablate this design choice in Appendix.

\subsection{Task Group Information}
We provide detailed task group information here. We follow  Table~\ref{tab:task_group} for multitask adaptation where group of 1 task means using "Target task" column only, group of 2 tasks means target task with task 1, and group of 3 tasks means target tasks, task 1, with task 2.

\begin{table}[t]
    \tabstyle{3pt}
    \footnotesize
    \caption{Task group for CoOp, VPT, and UPT
    }
    \label{tab:task_group}
    \begin{tabular}{l |c |c c}
    \toprule
    \multirow{2}{*}{Model} &  \multirow{2}{*}{Target task} & \multirow{2}{*}{Task 1} & \multirow{2}{*}{Task 2} \\
    & & & \\
    \midrule
   & Caltech101 & DTD & CIFAR10 \\
     & CIFAR10 & VOC 2007 & Resisc-45 \\
     & CIFAR100 & Caltech101 & CIFAR10 \\
     & Country-211 & Caltech101 & Resisc-45 \\
     & DTD & Caltech101 & MNIST \\
     & EuroSat & Resisc-45 & CIFAR100 \\
     & FER 2013 & CIFAR100 & MNIST \\
     & FGVCAircraft & Caltech101 & DTD \\
     & Flowers102 & CIFAR10 & Caltech101 \\
    CoOp   & Food101 & Caltech101 & DTD \\
     & GTSRB & MNIST & CIFAR100 \\
     & Hateful Memes & VOC 2007 & Caltech101 \\
     & KITTI Distance & StanfordCars & OxfordPets \\
     & MNIST & DTD & Resisc-45 \\
     & OxfordPets & Caltech101 & CIFAR10 \\
     & Patch-Camelyon & CIFAR100 & Caltech101 \\
     & Rendered-SST2 & FGVCAircraft & Hateful Memes \\
     & Resisc-45 & Caltech101 & CIFAR10 \\
     & StanfordCars & Caltech101 & MNIST \\
     & VOC 2007 & CIFAR100 & Caltech101 \\
    \midrule
     & Caltech101 & CIFAR100 & CIFAR10 \\
     & CIFAR10 & CIFAR100 & Caltech101 \\
     & CIFAR100 & CIFAR10 & Caltech101 \\
     & Country-211 & EuroSat & Food101 \\
     & DTD & CIFAR10 & Rendered-SST2 \\
     & EuroSat & Resisc-45 & FER 2013 \\
     & FER 2013 & OxfordPets & MNIST \\
     & FGVCAircraft & EuroSat & CIFAR10 \\
     & Flowers102 & CIFAR100 & EuroSat \\
    VPT & Food101 & CIFAR10 & EuroSat \\
     & GTSRB & MNIST & CIFAR100 \\
     & Hateful Memes & FER 2013 & OxfordPets \\
     & KITTI Distance & VOC 2007 & Flowers102 \\
     & MNIST & Resisc-45 & GTSRB \\
     & OxfordPets & CIFAR10 & Rendered-SST2 \\
     & Patch-Camelyon & CIFAR10 & Food101 \\
     & Rendered-SST2 & Resisc-45 & Patch-Camelyon \\
     & Resisc-45 & EuroSat & CIFAR10 \\
     & StanfordCars & CIFAR10 & EuroSat \\
     & VOC 2007 & CIFAR100 & CIFAR10 \\
    \midrule
     & Caltech101 & CIFAR10 & CIFAR100 \\
     & CIFAR10 & CIFAR100 & Caltech101 \\
     & CIFAR100 & Caltech101 & EuroSat \\
     & Country-211 & Caltech101 & CIFAR100 \\
     & DTD & CIFAR10 & Caltech101 \\
     & EuroSat & Resisc-45 & CIFAR100 \\
     & FER 2013 & MNIST & DTD \\
     & FGVCAircraft & CIFAR100 & CIFAR10 \\
     & Flowers102 & Caltech101 & CIFAR100 \\
   UPT  & Food101 & Caltech101 & CIFAR10 \\
     & GTSRB & MNIST & CIFAR100 \\
     & Hateful Memes & Caltech101 & CIFAR100 \\
     & KITTI Distance & Food101 & Flowers102 \\
     & MNIST & CIFAR100 & GTSRB \\
     & OxfordPets & Caltech101 & CIFAR100 \\
     & Patch-Camelyon & Food101 & StanfordCars \\
     & Rendered-SST2 & Hateful Memes & GTSRB \\
     & Resisc-45 & CIFAR10 & CIFAR100 \\
     & StanfordCars & Caltech101 & CIFAR100 \\
     & VOC 2007 & CIFAR100 & Caltech101 \\
    \bottomrule
    \end{tabular}
\end{table}

\end{document}